\documentclass[runningheads]{llncs}

\usepackage{eccv}

\usepackage{eccvabbrv}

\usepackage{graphicx}
\usepackage{booktabs}

\usepackage[accsupp]{axessibility}  

\usepackage[colorlinks,citecolor=eccvblue]{hyperref}

\usepackage{orcidlink}
\usepackage{multirow}
\usepackage{pifont} 
\usepackage{colortbl}
\usepackage{arydshln}
\usepackage{soul}
\usepackage{wrapfig}
\usepackage{bbm}
\usepackage{fontawesome}

\makeatletter
\renewcommand{\paragraph}{%
  \@startsection{paragraph}{4}%
  {\z@}{0.6ex \@plus .2ex \@minus .2ex}{-1em}
  {\normalfont\normalsize\bfseries}%
}
\makeatother

\def\modelname{ConceptExpress\xspace}

\definecolor{tokengreen}{HTML}{35A29F}
\definecolor{tokenred}{HTML}{EE7B7B}
\definecolor{tokenpurple}{HTML}{6A67CE}

\definecolor{concept1}{HTML}{362FD9}
\definecolor{concept2}{HTML}{176B87}
\definecolor{concept3}{HTML}{CD5C08}

\def\conceptone{\textcolor{concept1}{\ding{182}}\xspace}
\def\concepttwo{\textcolor{concept2}{\ding{183}}\xspace}
\def\conceptthree{\textcolor{concept3}{\ding{184}}\xspace}

\definecolor{segr}{HTML}{FFDDE3}
\definecolor{segy}{HTML}{FFE699}
\definecolor{fgr}{HTML}{C00000}
\definecolor{fgy}{HTML}{FF8C00}
\definecolor{segg}{HTML}{7F7F7F}

\def\markcheckone{\colorbox{fgr}{\textcolor{segr}{\faCheckCircle}}\xspace}
\def\markchecktwo{\colorbox{fgy}{\textcolor{segy}{\faCheckCircle}}\xspace}
\def\markquestion{\textcolor{segg}{\faQuestionCircle}\xspace}
\def\marktimes{\textcolor{segg}{\faTimesCircle}\xspace}

\def\conceptthree{\textcolor{concept3}{\ding{184}}\xspace}

\newcommand{\rvs}[1]{#1}
\newcommand{\rvsnew}[1]{#1}
\newcommand{\crhsz}[1]{#1}

\newcommand{\cmark}{\ding{51}}%

\begin{document}

\title{\modelname: Harnessing Diffusion Models for Single-image Unsupervised Concept Extraction} 

\titlerunning{\modelname}

\author{
Shaozhe Hao \and 
Kai Han\thanks{Corresponding authors} \and
Zhengyao Lv \\
Shihao Zhao \and
Kwan-Yee K. Wong$^\star$
}

\authorrunning{S.~Hao et al.}

\institute{
The University of Hong Kong\\
\email{\{szhao,shzhao,kykwong\}@cs.hku.hk kaihanx@hku.hk cszy98@gmail.com}
}

\maketitle

\begin{abstract}
While personalized text-to-image generation has enabled the learning of a single concept from multiple images, a more practical yet challenging scenario involves learning multiple concepts within a single image. However, existing works tackling this scenario heavily rely on extensive human annotations.
In this paper, we introduce a novel task named Unsupervised Concept Extraction (UCE) that considers an unsupervised setting without any human knowledge of the concepts.
Given an image that contains multiple concepts, the task aims to extract and recreate individual concepts solely relying on the existing knowledge from pretrained diffusion models. 
To achieve this, we present \modelname that tackles UCE by unleashing the inherent capabilities of pretrained diffusion models in two aspects. Specifically, a concept localization approach automatically locates and disentangles salient concepts by leveraging spatial correspondence from diffusion self-attention; and based on the lookup association between a concept and a conceptual token, a concept-wise optimization process learns discriminative tokens that represent each individual concept.
Finally, we establish an evaluation protocol tailored for the UCE task. Extensive experiments demonstrate that \modelname is a promising solution to the UCE task. Our code and data are available at: \url{https://github.com/haoosz/ConceptExpress}
\end{abstract}
\section{Introduction}
\label{sec:intro}
After observing an image containing multiple concepts, a skilled painter can recreate each individual concept within the complex scene. This remarkable cognitive ability prompts us to raise an intriguing question: \emph{Do text-to-image generative models also possess the capability to extract and recreate concepts?} In this paper, we try to provide an answer to this question by harnessing the potential of Stable Diffusion~\cite{rombach2022high} in concept extraction.

Diffusion models~\cite{ho2020denoising,songdenoising,ramesh2022hierarchical,nichol2022glide,saharia2022photorealistic,rombach2022high} have exhibited unprecedented performance in photorealistic text-to-image generation. Although diffusion models are trained solely for the purpose of text-to-image generation, extensive evidence suggests their underlying capabilities in various tasks, including classification~\cite{li2023diffusion}, segmentation~\cite{ni2023ref,karazija2023diffusion,tian2023diffuse,wang2023diffusion,xiao2023text}, and semantic correspondence~\cite{li2023sd4match,zhang2023tale,hedlin2023unsupervised}. This indicates that diffusion models embed significant world knowledge, potentially enabling them to perceive and recreate concepts akin to skilled painters. Motivated by this insight, we delve into this problem and explore the untapped potential of Stable Diffusion~\cite{rombach2022high} in concept extraction.
While recent research~\cite{avrahami2023break,jin2023image} has made initial attempts in exploring concept extraction using Stable Diffusion, existing approaches heavily rely on external human knowledge for supervision during the learning process.
For example, Break-A-Scene~\cite{avrahami2023break} demands pre-annotated object masks, while MCPL~\cite{jin2023image} requires accurate concept-descriptive captions. However, these human aids are both costly and often inaccessible. 
This critical constraint renders existing approaches infeasible, as none of them extract concepts without using any prior knowledge of the concepts.

\begin{figure}[t]
    \centering
    \includegraphics[width=0.99\linewidth]{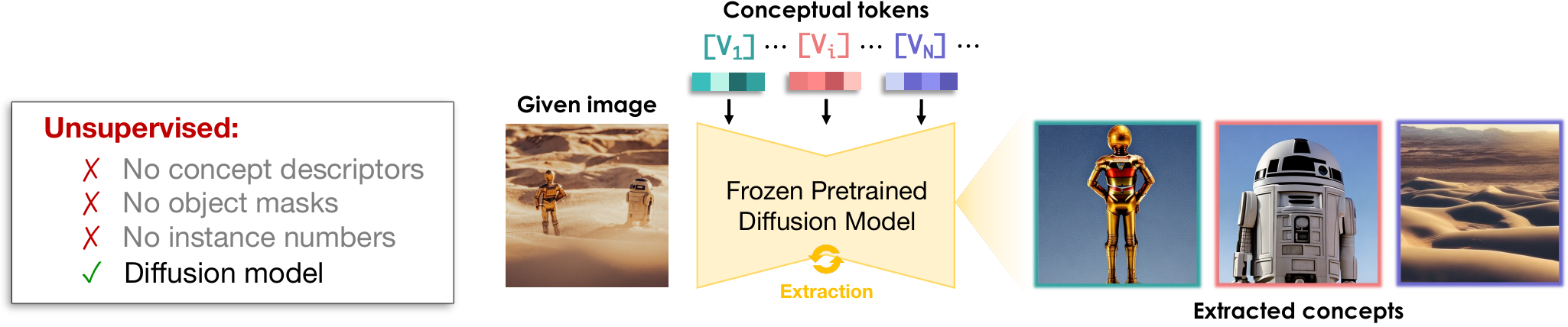}
    \caption{\textbf{Unsupervised concept extraction.} 
    We focus on the unsupervised problem of extracting multiple concepts from a single image. Given an image that contains multiple concepts (e.g., Star Wars characters \textcolor{tokengreen}{C-3PO}, \textcolor{tokenred}{R2-D2}, and \textcolor{tokenpurple}{desert}), we aim to harness a frozen pretrained diffusion model to automatically learn the conceptual tokens. Using the learned conceptual tokens, we can regenerate the extracted concepts with high quality, as shown in the rightmost column. In this process, no human knowledge or aids are available, and we only rely on the inherent capabilities of the pretrained Stable Diffusion~\cite{rombach2022high}.
    }
    \label{fig:teaser}
    \vspace{-5pt}
\end{figure}

To bridge this gap, we introduce a novel and challenging task named \textit{Unsupervised Concept Extraction} (UCE). Given an image containing multiple objects, UCE aims to automatically extract the object concepts such that they can be used to generate new images. 
In UCE, we consider a strict and realistic ``unsupervised'' setting, in which there is no prior knowledge about the image or the concepts present within it. Specifically, ``unsupervised'' emphasizes \textbf{(1)} no concept descriptors for proper word embedding initialization, \textbf{(2)} no object masks for concept localization and disentanglement, and \textbf{(3)} no instance number for a definite number of concepts to be extracted. 
We illustrate UCE in~\cref{fig:teaser}.

To tackle this problem, we introduce \textbf{\modelname}, the first method designed for unsupervised concept extraction. \modelname unleashes the inherent capabilities of pretrained Stable Diffusion, enabling it to disentangle each concept in the compositional scene and learn discriminative conceptual tokens that represent each individual concept. \modelname presents two major innovations.
(1)~For concept disentanglement, we propose a concept localization approach that automatically locates salient concepts within the image. This approach involves clustering spatial points on the self-attention map, building upon the observation that Stable Diffusion has learned good unsupervised spatial correspondence in the self-attention layers~\cite{tian2023diffuse}. Our approach has three sequential steps, namely pre-clustering, filtering, and post-clustering, seamlessly integrating a parameter-free hierarchical clustering method~\cite{sarfraz2019efficient}. By utilizing the end-of-text cross-attention map as a magnitude filter, we filter out non-salient backgrounds. Additionally, our approach automatically determines the number of concepts based on self-adaptive clustering constraints.
(2)~For conceptual token learning, we employ concept-wise masked denoising optimization by reconstructing the located concept. This optimization is based on a token lookup table that associates each located concept with its corresponding conceptual token. To address the issue of absence of initial words, which can detrimentally impact optimization~\cite{gal2022image}, we introduce a split-and-merge strategy for robust token initialization, mitigating performance degradation. To prevent undesired cross-attention activation with the wrong concept, we incorporate regularization to align cross-attention maps with the desired concept activation exhibited in self-attention maps.

To evaluate the new UCE task, we construct a new dataset that contains various multi-concept images, and introduce an evaluation protocol including two metrics tailored for unsupervised concept extraction. We use concept similarity, including identity similarity and compositional similarity, to measure the absolute similarity between the source and the generated concepts. We also use classification accuracy to assess the degree of concept disentanglement. Through comprehensive experiments, our results demonstrate that \modelname successfully tackles the challenge of unsupervised concept extraction, as evidenced by both qualitative and quantitative evaluations. 
\section{Related Work}
\label{sec:related}
\paragraph{Text-to-image synthesis}
In the realm of GANs~\cite{gan, brock2018large,karras2019style,karras2020analyzing,karras2021alias}, plenty of works have gained remarkable advancements in text-to-image generation~\cite{reed2016generative,Zhu_2019_CVPR,tao2022df,xu2018attngan,zhang2021cross,ye2021improving} and text-driven image manipulation~\cite{gal2022stylegan, patashnik2021styleclip,xia2021tedigan,abdal2022clip2stylegan}, significantly pushing forward image synthesis conditioned on plain text. Content-rich text-to-image generation is achieved by auto-regressive models~\cite{ramesh2021zero, yuscaling} that are trained on large-scale text-image data. Based on the pretrained CLIP~\cite{radford2021learning}, Crowson~\etal~\cite{crowson2022vqgan} optimizes the generated image at test time using CLIP similarity without any training. Diffusion-based methods~\cite{ho2020denoising} have pushed the boundaries of text-to-image generation to a new level, \eg, DALL·E~2~\cite{ramesh2022hierarchical}, Imagen~\cite{saharia2022photorealistic}, GLIDE~\cite{nichol2022glide}, and LDM~\cite{rombach2022high}.  
Based on the implementation of LDMs~\cite{rombach2022high}, Stable Diffusion (SD), large-scale trained on LAION-5B~\cite{schuhmann2022laion}, achieves unprecedented text-to-image synthesis performance.
Diffusion models are widely used for various tasks such as controllable generation~\cite{zhang2023adding,zhao2023uni}, global~\cite{brooks2022instructpix2pix,Tumanyan_2023_CVPR} and local editing~\cite{Avrahami_2022_CVPR,avrahami2023blended,couairon2022diffedit,kawar2023imagic,patashnik2023localizing,wang2023imagen}, video generation~\cite{hovideo,singer2022make,wu2022tune} and editing~\cite{molad2023dreamix,zhang2024controlvideo}, inpainting~\cite{lugmayr2022repaint}, and scene generation~\cite{avrahami2023spatext,bar2023multidiffusion}.

\paragraph{Generative concept learning}
Recently, many works~\cite{gal2022image,ruiz2022dreambooth,kumari2023multi,wei2023elite,jia2023taming,chen2023subject,qiu2023controlling,tewel2023keylocked,ma2023unified,shi2023instantbooth,li2023blip,gal2023designing,hao2023vico} have emerged, aiming to learn a generative concept from multiple images. For example, Textual Inversion~\cite{gal2022image} learns an embedding vector that represents a concept in the textual embedding space. Liu~\etal~\cite{liu2023unsupervised} extended it to multi-concept discovery using composable diffusion models~\cite{du2020compositional}. Their work operates in an unsupervised setting like ours. However, there is a major difference: they extract concepts from multiple images, each containing only one concept, whereas our focus is on extracting multiple concepts from a single image. Our work is closely related to Break-A-Scene~\cite{avrahami2023break} which relies heavily on human-annotated masks that are not available in our setting. \crhsz{The concurrent works, MCPL~\cite{jin2023image} and DisenDiff~\cite{zhang2024attention}, address a similar problem, but they require either a concept-descriptive text caption or specific class names, which renders them infeasible for our task.}
There are also works related to generative concepts include concept erasing~\cite{gandikota2023erasing,kumari2023ablating}, decomposition~\cite{vinker2023concept,chefer2023hidden}, manipulation~\cite{wang2023concept}, and creative generation~\cite{richardson2023conceptlab}. 

\paragraph{Attention-based segmentation}
Pre-trained Stable Diffusion~\cite{rombach2022high} possesses highly informative semantic representations within its attention layers. This property effectively enables its cross-attention layers to indicate the interrelations between text and image tokens~\cite{tang2022daam}, and its self-attention layers to capture the spatial correspondence among image tokens.
Consequently, prior works~\cite{baranchuklabel,ni2023ref,karazija2023diffusion,tian2023diffuse,wang2023diffusion,xiao2023text} have explored the utilization of the pre-trained Stable Diffusion for semantic segmentation, showing remarkable performance in unsupervised zero-shot segmentation. Diffsegmenter~\cite{wang2023diffusion} and FTM~\cite{xiao2023text} use cross-attention to initialize segmentation maps, and then extract affinity weights from self-attention for further refinement. DiffSeg~\cite{tian2023diffuse} achieves unsupervised zero-shot segmentation by clustering aggregated self-attention maps. Their investigation of the self-attention property inspires our concept localization approach.

\section{Unsupervised Concept Extraction}
\label{sec:method}

\begin{figure}[t]
    \centering
    \includegraphics[width=0.85\linewidth]{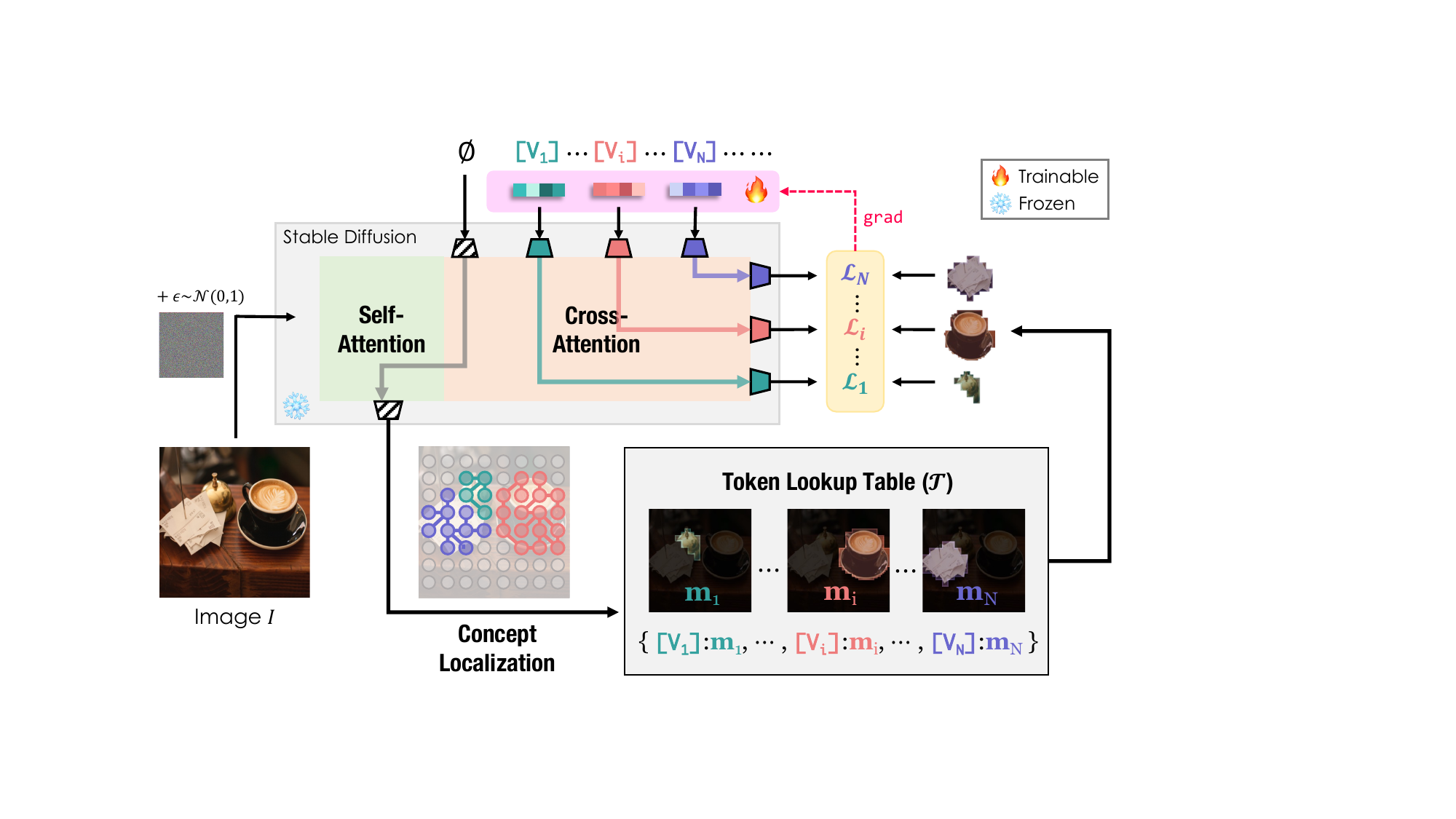}
    \caption{\textbf{Overview of \modelname.}
    \modelname takes a multi-concept image $\mathcal{I}$ as input and learns a set of conceptual tokens. \modelname consists of three key components. First, it leverages self-attention maps from the unconditional token $\varnothing$ to locate the latent concepts. Second, it constructs a token lookup table that associates each concept mask with its corresponding conceptual token $\mathtt{[V_i]}$. Finally, it optimizes each conceptual token using a masked denoising loss. The learned conceptual tokens can then be used to generate images that represent each individual concept. See~\cref{sec:method} for more details of the method.}
    \label{fig:pipe}
\end{figure}

We aim to learn discriminative tokens that can represent multiple instance-level concepts from a single image in an \emph{unsupervised} manner. Specifically, given an image $\mathcal{I}$ containing multiple salient instances, we use a pretrained text-to-image diffusion model to discover a set of conceptual tokens $\mathcal{S} = \{\mathtt{[V_i]}\}_{i=1}^{N}$ and their corresponding embedding vectors $\mathcal{V}=\{v_i\}_{i=1}^{N}$, which capture discriminative concepts from $\mathcal{I}$. The concept number $N$ is automatically determined in the discovery process. By prompting the $i$-th token $\mathtt{[V_i]} \in \mathcal{S}$, we can recreate the corresponding concept extracted from $\mathcal{I}$. We present \modelname to tackle this problem. \cref{fig:pipe} gives an overview of \modelname.

\subsection{Preliminary} 
\textbf{Text-to-image diffusion model}~\cite{rombach2022high} is composed of a pretrained autoencoder with an encoder $\mathcal{E}$ to extract latent codes and a corresponding decoder $\mathcal{D}$ to reconstruct images, a CLIP~\cite{radford2021learning} text encoder that extracts text embeddings, and a denoising U-Net $\epsilon_\theta$ with text-conditional cross-attention blocks. Textual inversion~\cite{gal2022image} represents a particular concept using a learnable embedding vector $v_\star$, which is optimized using a standard latent denoising loss with $\epsilon_\theta$ frozen, written as
\begin{equation}
\setlength{\abovedisplayskip}{5pt}
\setlength{\belowdisplayskip}{5pt}
    \mathcal{L}=\mathbb{E}_{z\sim\mathcal{E}(\mathcal{I}), y, \epsilon\sim\mathcal{N}(0,1), t}\left[\Vert \epsilon - \epsilon_\theta(z_t, t, c_{v_\star}(y)) \Vert^2_2\right],
\end{equation}
where $t$ is the timestep, $z_t$ is the latent code at timestep $t$, $\epsilon$ is the randomly sampled Gaussian noise, $y$ is the text prompt, and $c_{v_\star}$ is the text encoder parameterized by the learnable $v_\star$. 
\modelname advances further by learning multiple embedding vectors in an unsupervised setting.\vspace{5pt}\\
\rvs{\textbf{FINCH}~\cite{sarfraz2019efficient} is an efficient parameter-free hierarchical clustering method. Given a set of $n$ sample points in $d$ dimensions, denoted as $\mathcal{S} = \{s_i\ |\ s_i \in \mathbb{R}^{d} \}_{i=1}^{n}$, we construct an adjacent matrix $\mathrm{G}$ for paired samples as
\begin{equation}
\setlength{\abovedisplayskip}{2pt}
\setlength{\belowdisplayskip}{2pt}
    \mathrm{G}(i,j)=\begin{cases}
    1& \text{if} \  \kappa_i=j \ \text{or} \  \kappa_j=i \ \text{or} \ \kappa_i=\kappa_j \\
    0& \text{otherwise}
    \end{cases},
    \label{eq:adjacency}
\end{equation}
where $\kappa_i$ represents the index of the closest sample to $s_i \ \in\ \mathcal{S}$ under a specific distance metric. 
To obtain a sample partition, we group the connected components within the undirected graph defined by the adjacency matrix $\mathrm{G}$. Each connected component in the graph represents a cluster, and the centroids of the clusters are treated as super sample points for constructing a new adjacent matrix. 
This process enables iterative hierarchical clustering until all samples are grouped. As a result, multiple clustering levels of varying granularity are generated.
}

\subsection{Automatic Latent Concept Localization}
\label{sec:latent-mask}
We begin by locating instance-level concepts within the diffusion latent space. In pretrained diffusion models, self-attention possesses good properties of spatial correspondence which offers the inherent benefit as an unsupervised semantic segmenter~\cite{tian2023diffuse}. With this insight, we propose an approach to automatically locating concepts by subtly leveraging self-attention. 

\begin{figure}[t]
    \centering
    \includegraphics[width=\linewidth]{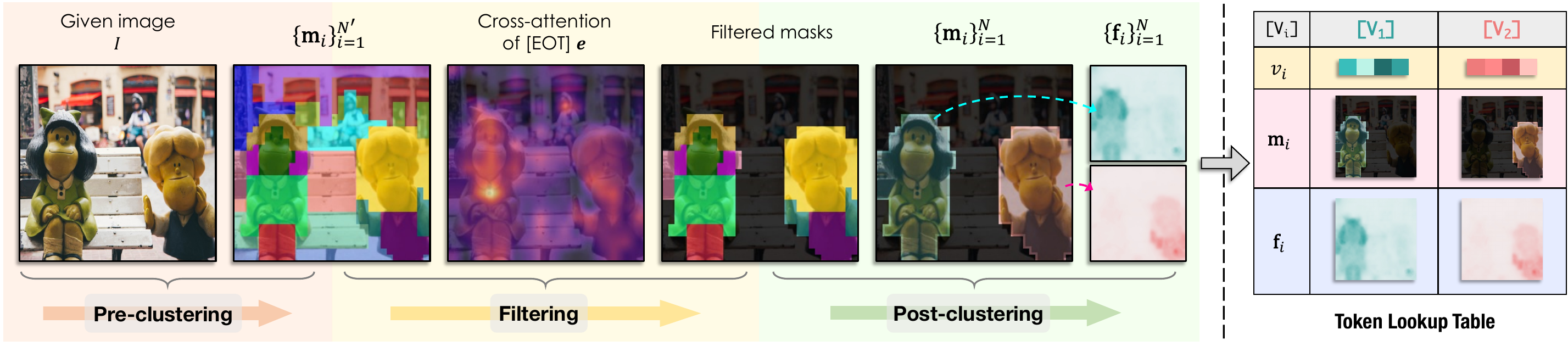}
    \caption{\textbf{Visualization.} \textbf{Left}: we visualize the concept localization process, which involves: (1) pre-clustering that groups together semantically related regions; (2) filtering that removes non-salient regions that are not visually significant; and (3) post-clustering that integrates salient regions into instance-level concepts. \textbf{Right}: we visualize the token lookup table, which establishes a one-to-one correspondence between the conceptual token $\mathtt{[V_i]}$ and the learnable embedding vector $v_i$, the latent mask $\mathbf{m}_i$, and the attention map $\mathbf{f}_i$.
    }
    \label{fig:concept-localization}
    \vspace{-1pt}
\end{figure}

Let $\mathbf{A}_l \in \mathbb{R}^{(h_l \times w_l) \times (h_l \times w_l)}$ denote the self-attention map from the $l$-th layer of the U-Net, where the feature map has a spatial resolution $h_l \times w_l$. To aggregate self-attention maps from different layers into an identical resolution $h \times w$, we follow the practice in~\cite{tian2023diffuse} to interpolate the last two dimensions, duplicate the first two dimensions, and average all maps. The aggregated attention, denoted as $\mathbf{A} \in \mathbb{R}^{(h \times w) \times (h \times w)}$, can be represented as a set of $h \times w$ spatial samples, each of which is an $h \times w$ dimensional distribution, \ie, $\mathcal{A} = \{a_i\ |\ a_i \in \mathbb{R}^{h \times w} \}_{i=1}^{h\times w}$. 
By clustering on $\mathcal{A}$, we can naturally derive latent masks that align with the semantic segmentation of the original image. This is because latent patches sharing similar semantics tend to possess consistent self-attention activations. The masks are formed by combining spatial samples belonging to the same cluster, effectively representing specific segments in the image. However, accurately locating instance-level concepts and effectively filtering out the background remain challenging when our goal is to disentangle multiple instances rather than solely segmenting semantics. To tackle this challenge, we adapt the hierarchical clustering algorithm FINCH~\cite{sarfraz2019efficient} to generate latent masks that satisfy our needs.

\paragraph{Pre-clustering}  \rvs{We first apply FINCH algorithm on $\mathcal{A}$. 
Since $a_i$ is normalized and treated as a distribution, $\kappa_i$ can be determined using a distribution distance metric, specifically the mean KL divergence, \ie,}
\begin{align}
\setlength{\abovedisplayskip}{0pt}
\setlength{\belowdisplayskip}{0pt}
    d(a_i, a_j) &= \left(D_{KL}\left(a_i, a_j\right) + D_{KL}\left(a_j, a_i\right)\right)/2,  \\
    \kappa_i &= \mathop{\arg\min}_{j} \ \{d(a_i, a_j) \ |\ a_j \in \mathcal{A}\}.
    \label{eq:neighbor}
\end{align}
We set the upper limit of the number of discovered concepts to $N_{\text{max}}$. We then identify the clustering level with the cluster number $N^\prime$ closest to but greater than $N_{\text{max}}$. At this level, we construct a mask for each cluster from all spatial points within the cluster. We denote the resulting masks as $\{\mathbf{m}_i\ |\ \mathbf{m}_i \in \{0, 1\}^{h \times w}\}_{i=1}^{N^\prime}$. Since spatial samples within the same cluster share consistent semantics, the distribution distance between them serves as an effective indicator for distinguishing between different semantic instances. 
\rvsnew{Therefore, we use the largest intra-cluster distance at this level, denoted as $\delta$, as a self-adaptive threshold to determine the final clustering level in the post-clustering phase.}

\paragraph{Filtering} The obtained masks cover all areas on the latent map, encompassing both the foreground instances with clear semantics and the indistinct background regions. In diffusion models, the cross-attention map of the end-of-text token (\texttt{[EOT]}) demonstrates robust foreground localization capabilities~\cite{hao2023vico}, where salient regions exhibit higher magnitudes and vice versa. This characteristic makes it suited for automatically distinguishing between distinct instances and indistinct backgrounds. Let $\mathbf{e} \in \mathbb{R}^{h \times w}$ denote the cross-attention map of \texttt{[EOT]}. Based on $\mathbf{e}$, we discard masks whose masked regions satisfy
\begin{equation}
\label{eq:filtering}
\setlength{\abovedisplayskip}{5pt}
\setlength{\belowdisplayskip}{5pt}
    \frac{\Vert {\rm vec}(\mathbf{m}_i \odot \mathbf{e}) \Vert_1}{\Vert {\rm vec}(\mathbf{m}_i) \Vert_1} < \frac{\Vert {\rm vec}(\mathbf{e}) \Vert_1}{h\times w}
\end{equation}
where ${\rm vec}(\cdot)$, $\Vert \cdot \Vert_1$, and $\odot$ denote matrix vectorization, $\ell_1$ norm, and Hadamard product respectively. By applying this criterion, we filter out those masks whose masked regions show magnitudes lower than the average level, indicating that they correspond to indistinct regions.
This criterion helps identify and exclude masks that correspond to indistinct regions in the \texttt{[EOT]} cross-attention map. 
\paragraph{Post-clustering} \rvs{After filtering, we reapply FINCH to the remaining clusters iteratively.} Additionally, we introduce two extra constraints to determine the stopping point in the clustering procedure. 
\textbf{(1)}~To enhance the proximity of semantic relationships within the same mask, we set $\mathrm{G}(i,j) \tiny{=} 0$ if the distance $d(a_i, a_j)$ exceeds $\delta$, which is determined in the level with $N^\prime$ clusters of pre-clustering. By removing such connections, we hinder the grouping of strong semantic variations within the same mask.
\textbf{(2)}~We forbid non-adjacent masks from grouping together, \ie, masks that are not spatially adjacent to each other cannot be clustered together, regardless of their connectivity in \textrm{G}. 
With these two constraints, the clustering will automatically terminate and yield $N$ masks that locate the latent spaces corresponding to the $N$ target concepts. The mean attention activations of each concept region is precisely the centroid of each cluster, given by
\begin{equation}
    \mathbf{f}_i = \mathbf{m}_i^{1 \times (h \times w)} \cdot \mathbf{A}^{(h \times w) \times (h \times w)}\ /\  \Vert {\rm vec}(\mathbf{m}_i) \Vert_1
\end{equation}
where the centroid $\mathbf{f}_i \in \mathbb{R}^{1 \times (h \times w)}$ represents the average attention activations of $i$-th masked latent region to the entire $h \tiny{\times} w$ latent space. The latent masks and their corresponding attention activations are ready for token optimization. The concept localization process is visualized in~\cref{fig:concept-localization}~(left).
\vspace{-1pt}

\subsection{Concept-wise Masked Denoising}
We construct a token lookup table
\begin{equation}
\setlength{\abovedisplayskip}{5pt}
\setlength{\belowdisplayskip}{5pt}
    \mathcal{T}_\text{lookup} := \left\{\mathtt{[V_i]}:(v_i, \mathbf{m}_i, \mathbf{f}_i)\ |\ i = 1, 2, \cdots, N \right\}
    \label{eq:lookup}
\end{equation}
where the $i$-th conceptual token $\mathtt{[V_i]}$ corresponds to a learnable embedding vector $v_i$, a latent mask $\mathbf{m}_i \in \{0,1\}^{h \times w}$, and a mean attention map $\mathbf{f}_i \in \mathbb{R}^{h \times w}$. We visualize the token lookup table in~\cref{fig:concept-localization}~(right). We employ the masked denoising loss~\cite{avrahami2023break} to optimize each token $\mathtt{[V_i]} \in  \mathcal{T}_\text{lookup}$:
\begin{equation}
\setlength{\abovedisplayskip}{5pt}
\setlength{\belowdisplayskip}{5pt}
    \mathcal{L}_i=\mathbb{E}_{z\sim\mathcal{E}(\mathcal{I}), y_i, \epsilon, t}\left[\left\Vert \left[\epsilon - \epsilon_\theta\left(z_t, t, c_{\textcolor{teal}{v_i}}(y_i)\right)\right] \odot \mathbf{m}_i \right\Vert^2_2\right]
    \label{eq:masked-loss}
\end{equation}
where $y_i$ is the text prompt ``\texttt{a photo of $\mathtt{[V_i]}$}" and \textcolor{teal}{$v_i$} is the only trainable parameter. Masked denoising forces the new token to learn exclusively within specific latent regions that contain concept-wise information.

\begin{wrapfigure}{r}{0.5\textwidth}
\vspace{-6pt}
    \centering
    \includegraphics[width=\linewidth]{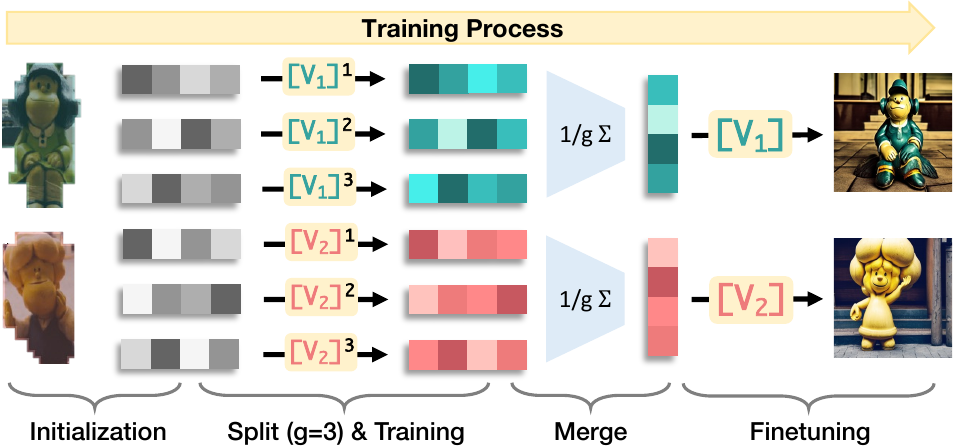}
    \caption{\textbf{Split-and-merge}. During the training process, we sequentially initialize conceptual tokens, train the split tokens, merge the tokens by averaging, and further fine-tune the merged tokens. Finally, the merged tokens are well-learned and effectively represent individual concepts.}
    \label{fig:split-and-merge}
    \vspace{-5pt}
\end{wrapfigure}

\label{sec:masked-denoising}
\paragraph{Robust token initialization}
To learn the concept embedding above, one may think to directly apply textual inversion~\cite{gal2022image}. However, the use of suitable words for initializing the conceptual tokens is crucial for successful textual inversion. In the case of an unsupervised setting, where specific words for initializing concept tokens are not available, the performance of~\cite{gal2022image} will deteriorate sharply.
To resolve this problem, we propose a \emph{split-and-merge} strategy that randomly initializes multiple tokens for each concept, which are later merged into a single token after several warm-up steps. Multiple tokens can explore a broader concept space, providing a greater opportunity for convergence into an embedding vector that can more precisely represent the underlying concept. Formally, we randomly initialize $g$ tokens $\{\mathtt{[{V_i}]^j}\}_{j=1}^{g}$ for each concept and extend the token lookup table as
\begin{equation}
\setlength{\abovedisplayskip}{5pt}
\setlength{\belowdisplayskip}{5pt}
    \mathcal{T}_\text{lookup}^{\text{split}} := \left\{\mathtt{[{V_i}]^j}:(v_i^j, \mathbf{m}_i, \mathbf{f}_i)\ |\ i \tiny{=} 1,...,N; j\tiny{=} 1,...,g\right\},
\end{equation}
where $v_i^j$ is the $j$-th randomly initialized embedding vector corresponding to the conceptual token $\mathtt{[V_i]^j}$. At the early training steps, we optimize the loss in~\cref{eq:masked-loss} on $\mathtt{[V_i]^j} \in \mathcal{T}_\text{lookup}^{\text{split}}$, \ie, $\mathcal{L}_{i,j}$, to learn the $g \times N$ tokens. In addition, leveraging the constraint that embeddings for the same concept should exhibit a closer embedding distance, we incorporate a contrastive loss for each token $\mathtt{[V_i]^j}$ as
\begin{equation}
\setlength{\abovedisplayskip}{5pt}
\setlength{\belowdisplayskip}{5pt}
\mathcal{L}^{con}_{i,j} = - \frac{1}{g\tiny{\times} N}\  log\frac{\sum_{v_i^q \in \mathcal{V}_i\setminus\{v_i^j\}}\exp(v_i^j \cdot v_i^q/\tau)}{\sum_{v_m^n \in \mathcal{V}\setminus\{v_i^j\}}\exp(v_i^j \cdot v_m^n/\tau)},
\label{eq:contrast}
\end{equation}
\crhsz{where $\tau$ is the temperature, $\mathcal{V}$ is the full set of embedding vectors, and $\mathcal{V}_i$ is the subset of embedding vectors that correspond to the $i$-th concept.}
\cref{eq:contrast} enforces tokens representing the same concept to be closer to each other, inducing these randomly initialized embedding vectors to converge to a shared space during several warm-up training steps. Afterward, we merge the tokens by computing the mean value of the $g$ embeddings associated with each concept. The token lookup table is reset to \cref{eq:lookup}, where the token embeddings are good initializers to robustly represent the corresponding concepts. In the subsequent training steps, we use the denoising loss described in~\cref{eq:masked-loss} to optimize the merged tokens that represent each concept on a one-to-one basis. We depict the training process using split-and-merge in~\cref{fig:split-and-merge}.

\paragraph{Attention alignment}
Although each conceptual token is optimized to reconstruct the masked region, there is a lack of direct alignment between the tokens and individual concepts within a compositional scene. This absence of alignment leads to inaccurate cross-attention activation for the learned conceptual tokens, which hinders the performance of compositional generation.
To address this problem, for each token in the lookup table, we align its cross-attention map with the mean attention $\mathbf{f}_i$ of the corresponding masked region using a location-aware earth mover's distance (EMD) regularization. The earth moving cost is computed as the Euclidean distance between the 2D locations on two attention maps. 
Let the cross attention map of the token $\mathtt{[V_i]^j}$ be $\mathbf{c}_{\mathtt{[V_i]^j}} \in \mathbb{R}^{h \times w}$, where $j$ can be omitted after token merging. The regularization loss is formulated as
\begin{equation}
\setlength{\abovedisplayskip}{3pt}
\setlength{\belowdisplayskip}{3pt}
    \mathcal{L}^{reg}_{i,j} = \mathtt{EMD}(\mathbf{c}_\mathtt{[V_i]^j},\mathbf{f}_i)
\end{equation}
which softly guides the cross-attention map to match the desired concept activations exhibited in the self-attention map.
\vspace{-10pt}
\subsection{Implementation Details}
\label{sec:implement}
We train the tokens in two phases for a total of 500 steps, with a learning rate of $\text{5e-4}$. In the first 100 steps, we optimize the tokens $\mathtt{[V_i]^j} \in \mathcal{T}_\text{lookup}^{\text{split}}$ using
\begin{equation}
\setlength{\abovedisplayskip}{3pt}
\setlength{\belowdisplayskip}{3pt}
    \mathcal{L}= \frac{1}{g\tiny{\times} N} \sum_{i=1}^N \sum_{j=1}^g (\mathcal{L}_{i,j} + \alpha \mathcal{L}^{con}_{i,j} + \beta \mathcal{L}^{reg}_{i,j}).
\end{equation}
We then merge the tokens, deriving $\mathtt{[V_i]} \in \mathcal{T}_\text{lookup}$, and optimize them in the subsequent 400 steps using
\begin{equation}
\setlength{\abovedisplayskip}{3pt}
\setlength{\belowdisplayskip}{3pt}
    \mathcal{L} = \frac{1}{N} \sum_{i=1}^N (\mathcal{L}_i + \beta \mathcal{L}^{reg}_i).
\end{equation}
We use Stable Diffusion v2-1~\cite{rombach2022high} as our base model. We set $\alpha$=1e-3, $\beta$=1e-5, $\tau$=0.07, and g=5. All experiments are conducted on a single RTX 3090 GPU. In our implementation, self-attention used in concept localization is computed using the unconditional text prompt $\varnothing$ at timestep~$0$, which induces minimal textual intervention and maximal denoising of the given image.
\vspace{-6pt}
\section{Experiments}
\subsection{Dataset and Baseline}
\paragraph{Dataset}
In our work, we do not rely on predefined object masks or manually selected initial words for training images. This allows us to gather high-quality images from the Internet without human annotations to form our dataset. Specifically, we collect a set $D_1$ of 96\footnote{\crhsz{96 is considerably large compared to the dataset sizes in the previous works, such as 30 in DreamBooth~\cite{ruiz2022dreambooth}, 50 in Break-A-Scene~\cite{avrahami2023break}, and 10 in DisenDiff~\cite{zhang2024attention}.}} images from Unsplash\footnote{\url{https://unsplash.com/}}, ensuring that each image contains at least two distinct instance-level concepts.  The collected images encompass a wide range of object categories, including animals, characters, toys, accessories, containers, sculptures, buildings, landscapes, vehicles, foods, and plants. 
For a fair comparison, we construct a set $D_2$ using the 7 images provided by~\cite{avrahami2023break}. For evaluation, we generate 8 testing images for each training image using the prompt ``a photo of $\mathtt{[V_i]}$''.

\paragraph{Baseline}
To the best of our knowledge, Break-A-Scene~\cite{avrahami2023break} is the only work that is closely related to the problem in this paper. However, Break-A-Scene operates under a strongly supervised setting, which requires significant prior knowledge of the training image, including the number of concepts, object masks, and properly selected initial words. To ensure a fair and meaningful comparison, we adapt Break-A-Scene to our unsupervised setting. 
Specifically, we disable the use of manually picked initial words and instead apply random initialization. 
Additionally, we leverage the instance masks identified by our method as the annotated masks for Break-A-Scene.
Finally, we exclusively train the learnable tokens without fine-tuning the diffusion model. We use this adapted version of Break-A-Scene, denoted as BaS$^\dag$, as the baseline method for comparison.

\subsection{Evaluation Metric}
We establish an evaluation protocol for the UCE problem, which includes two tailored metrics described as follows.
\paragraph{Concept similarity} To quantify how well the model is able to recreate the concepts accurately, we evaluate the concept similarity, including identity similarity (\textit{SIM}$^{\text{I}}$) and compositional similarity (\textit{SIM}$^\text{C}$). Identity similarity measures the similarity between each concept in the training image and the concept-specific generated images. We employ CLIP~\cite{radford2021learning} and DINO~\cite{caron2021emerging} to compute the similarities. 
\rvs{To ensure that the similarity is computed specifically for the $i$-th concept, we obtain concept-wise masks with SAM~\cite{kirillov2023segment} by identifying the specific SAM mask associated with our extracted concept. Specifically, for each concept, we prompt SAM with 3 randomly sampled points on our extracted mask to produce SAM masks. The training image is then masked with the SAM mask corresponding to the $i$-th concept.}
The metric of identity similarity provides a crucial criterion for evaluating the intra-concept performance of unsupervised concept extraction. Compositional similarity measures the CLIP or DINO similarity between the source image and the generated image, conditioned on the prompt ``a photo of $\mathtt{[V_1]}$ and $\mathtt{[V_2]}$ ... $\mathtt{[V_N]}$''. This metric quantifies the degree to which the source image can be reversed using the extracted concepts.

\paragraph{Classification accuracy} 
To assess the extent of disentanglement achieved for each concept within the full set of extracted concepts, we establish a benchmark that evaluates concept classification accuracy. Specifically, we first employ a vision encoder, such as CLIP~\cite{radford2021learning} or DINO~\cite{caron2021emerging}, to extract feature representations for each concept from the SAM-masked training images. In total, we obtain 264 concepts in $D_1$ and 19 concepts in $D_2$.
We use these concept features as prototypes to construct a concept classifier. We then employ the same vision encoder to extract query features for all generated images, each associated with a specific concept category. Finally, we evaluate the top-$k$ classification accuracy of the query features using our concept classifier. We report classification results for $k\tiny{=}1,3$, denoting the top-$k$ accuracy as \textit{ACC}$^k$. This metric effectively assesses the inter-concept performance of unsupervised concept extraction.

\vspace{-10pt}
\subsection{Performance}
\paragraph{Quantitative comparison}
We compare \modelname with BaS$^\dag$ based on concept similarity and classification accuracy metrics. The quantitative comparison results on the two datasets are reported in~\cref{tab:quant-clip,tab:quant-dino}, respectively with CLIP~\cite{radford2021learning} and DINO~\cite{caron2021emerging} as the visual encoder. Notably, \modelname outperforms BaS$^\dag$ by a significant margin on all evaluation metrics. It achieves higher concept similarity \textit{SIM}$^\text{I}$ and \textit{SIM}$^\text{C}$, indicating a closer alignment with the source concepts. It also achieves higher classification accuracy \textit{ACC}$^1$ and \textit{ACC}$^3$, indicating a more significant level of disentanglement among the individually extracted concepts. These results highlight the limitations of the existing concept extraction approach~\cite{avrahami2023break} and establish \modelname as the state-of-the-art method for the UCE problem.
\begin{figure}[t]
    \centering
    \includegraphics[width=\linewidth]{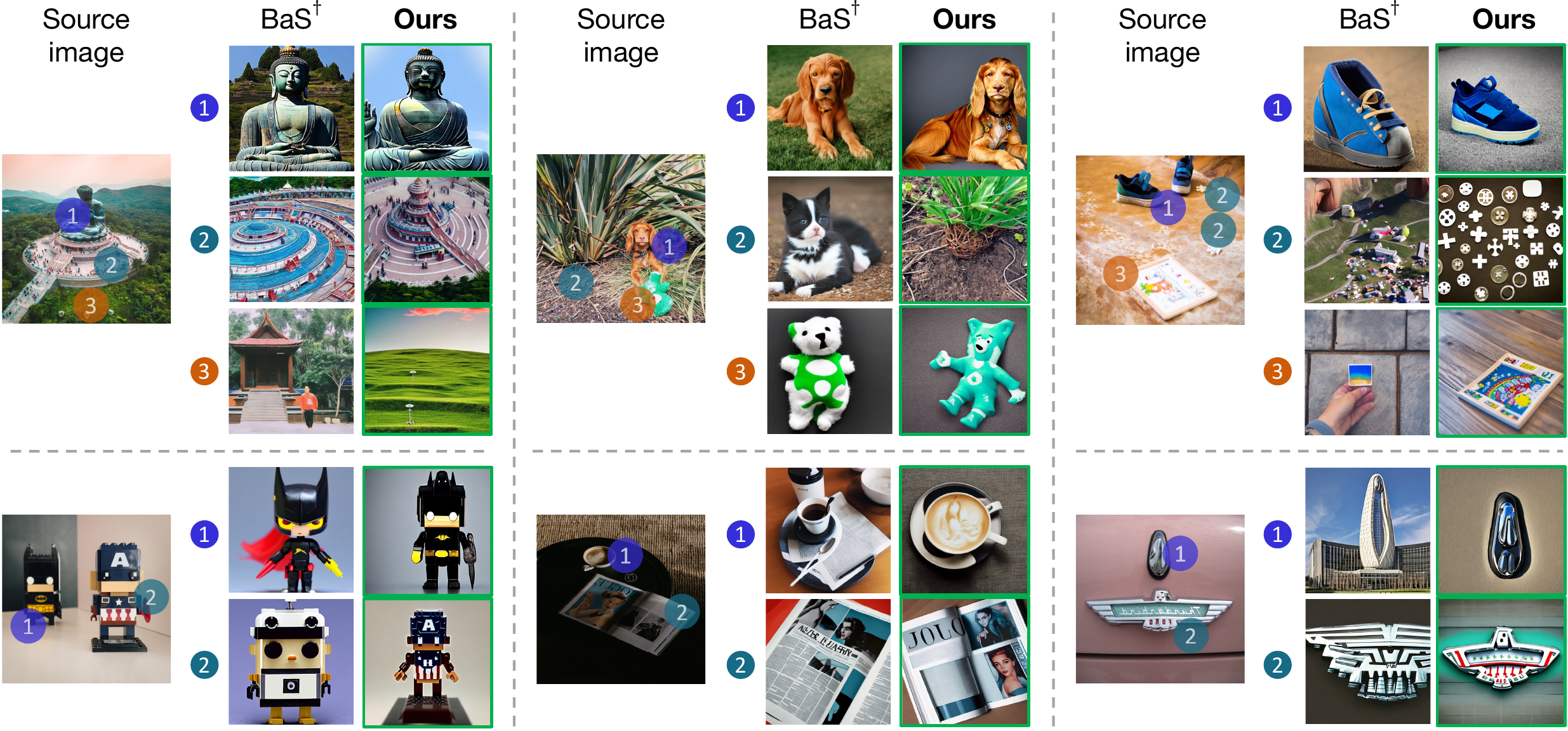}
    \caption{\textbf{Comparison with BaS$^\dag$~\cite{avrahami2023break}.} We compare the concept extraction results of BaS$^\dag$ and \modelname in 6 examples. For each example, we show the source image and the generated concept images. We annotate concepts in serial numbers for legibility. }
    \label{fig:main-compare}
\end{figure}

\begin{table}[t]
\caption{\textbf{\crhsz{Quantitative comparison.}} \crhsz{For reference, we also provide the results of original Break-A-Scene~\cite{avrahami2023break} on $D_2$ (marked in \textcolor{Gray}{grey}), by using mask and initializer supervision (BaS) and further finetuning (BaS f.t.).}\vspace{-13pt}}
\begin{subtable}[b]{0.49\linewidth}
\caption{Evaluation using CLIP~\cite{radford2021learning}.\vspace{-10pt}}
  \label{tab:quant-clip}
  \centering
  \scalebox{0.63}{
  \begin{tabular}{lcccccccc}
    \toprule
    & \multicolumn{4}{c}{$D_1$} & \multicolumn{4}{c}{$D_2$} \\
    \cmidrule(lr){2-5} \cmidrule(lr){6-9}
    Method & \textit{SIM}$^\text{I}$ & \textit{SIM}$^\text{C}$ & \textit{ACC}$^1$ & \textit{ACC}$^3$ & \textit{SIM}$^\text{I}$ & \textit{SIM}$^\text{C}$ &
    \textit{ACC}$^1$ & \textit{ACC}$^3$ \\
    \midrule
    \textcolor{Gray}{BaS}~\cite{avrahami2023break} & \textcolor{Gray}{--} & \textcolor{Gray}{--} & \textcolor{Gray}{--} & \textcolor{Gray}{--} & \textcolor{Gray}{0.686} & \textcolor{Gray}{0.696} &  \textcolor{Gray}{0.467} & \textcolor{Gray}{0.599} \\
    \textcolor{Gray}{BaS f.t.~\cite{avrahami2023break}} & \textcolor{Gray}{--} & \textcolor{Gray}{--} & \textcolor{Gray}{--} & \textcolor{Gray}{--} & \textcolor{Gray}{0.693} & \textcolor{Gray}{0.789} & \textcolor{Gray}{0.526} & \textcolor{Gray}{0.697} \\
    \arrayrulecolor{gray} \hdashline
    BaS$^\dag$~\cite{avrahami2023break} & 0.627 & 0.773 & 0.174 & 0.282 & 0.613 & 0.653 & 0.368 & 0.487 \\
    Ours & \textbf{0.689} & \textbf{0.784} & \textbf{0.263} & \textbf{0.385} & \textbf{0.715} & \textbf{0.737} & \textbf{0.566} & \textbf{0.783} \\
    \bottomrule
  \end{tabular}}
\end{subtable}
\begin{subtable}[b]{0.49\linewidth}
\caption{Evaluation using DINO~\cite{caron2021emerging}.\vspace{-10pt}}
  \label{tab:quant-dino}
  \centering
  \scalebox{0.63}{
  \begin{tabular}{lcccccccc}
    \toprule
    & \multicolumn{4}{c}{$D_1$} & \multicolumn{4}{c}{$D_2$} \\
    \cmidrule(lr){2-5} \cmidrule(lr){6-9}
    Method & \textit{SIM}$^\text{I}$ & \textit{SIM}$^\text{C}$ & \textit{ACC}$^1$ & \textit{ACC}$^3$ & \textit{SIM}$^\text{I}$ & \textit{SIM}$^\text{C}$ &
    \textit{ACC}$^1$ & \textit{ACC}$^3$ \\
    \midrule
    \textcolor{Gray}{BaS}~\cite{avrahami2023break} & \textcolor{Gray}{--} & \textcolor{Gray}{--} & \textcolor{Gray}{--} & \textcolor{Gray}{--} & \textcolor{Gray}{0.316} & \textcolor{Gray}{0.474} &  \textcolor{Gray}{0.559} & \textcolor{Gray}{0.704} \\
    \textcolor{Gray}{BaS f.t.~\cite{avrahami2023break}} & \textcolor{Gray}{--} & \textcolor{Gray}{--} & \textcolor{Gray}{--} & \textcolor{Gray}{--} & \textcolor{Gray}{0.411} & \textcolor{Gray}{0.696} & \textcolor{Gray}{0.697} & \textcolor{Gray}{0.737} \\
    \arrayrulecolor{gray} \hdashline
    BaS$^\dag$~\cite{avrahami2023break} & 0.254 & 0.510 & 0.202 & 0.315 & 0.231 & 0.417 & 0.329 & 0.559 \\
     Ours & \textbf{0.319} & \textbf{0.568} & \textbf{0.324} & \textbf{0.470} & \textbf{0.371} & \textbf{0.535} & \textbf{0.803} & \textbf{0.934} \\
    \bottomrule
  \end{tabular}}
\end{subtable}
\end{table}

\paragraph{Qualitative comparison}
We show several generation samples of \modelname and BaS$^\dag$ in~\cref{fig:main-compare}. \modelname presents overall better generation fidelity and quality than BaS$^\dag$. 
We observe some defects in the generations of BaS$^\dag$. For example, in the top left \conceptthree and the top center \concepttwo, the generation of BaS$^\dag$ deviates from the source concept. In addition, BaS$^\dag$ fails to preserve the characteristics of the source concept in the top center \conceptthree and the bottom left \conceptone\concepttwo.
\modelname effectively overcomes the defects of wrong identity and poor preservation observed in the generations in BaS$^\dag$. \modelname consistently generates high-quality images that precisely align with the source concepts.

\subsection{Ablation Study}
\begin{figure}[t]
\begin{minipage}[b]{0.54\linewidth}
    \centering
    \includegraphics[width=\linewidth]{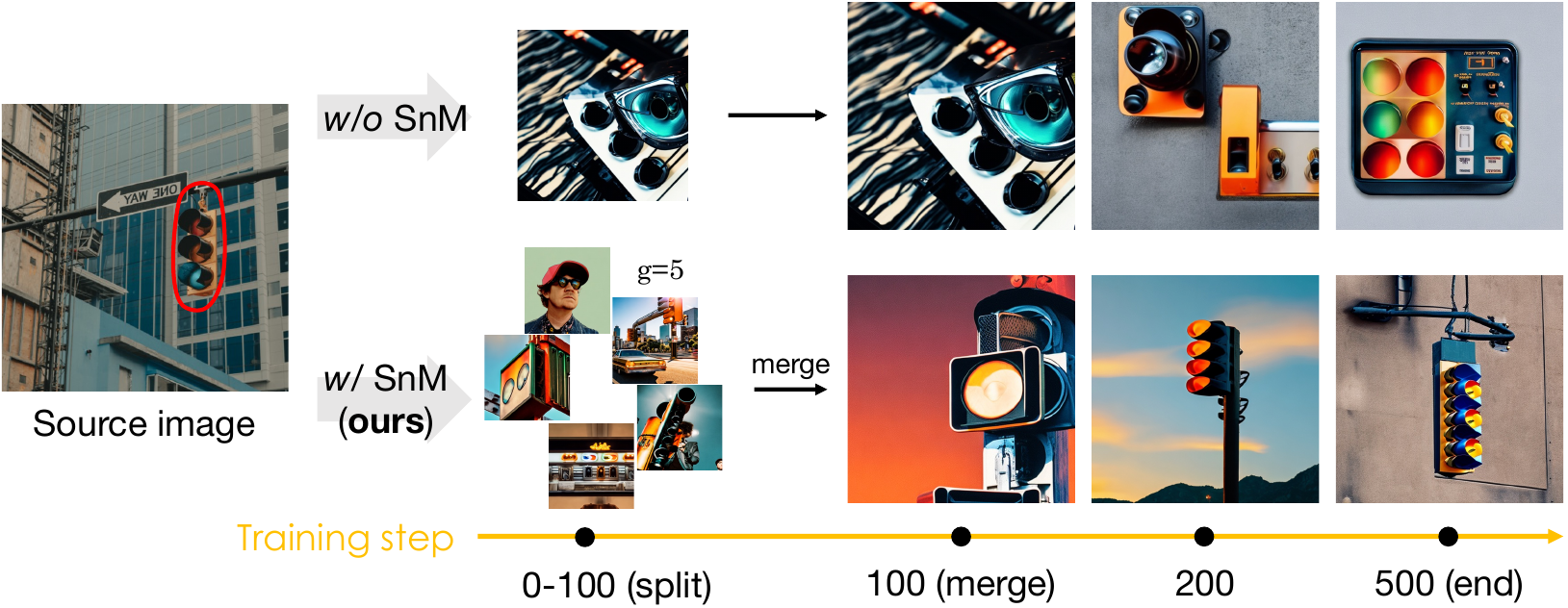}
    \caption{\textbf{Generation results of Split-and-merge (SnM) ablation.} We show the generated concept ``traffic light'' throughout the training process, with (bottom) and without (top) SnM that utilizes $g\tiny{=5}$ diverse tokens.
    }
    \label{fig:ab-split}
\end{minipage}\ \ 
\begin{minipage}[b]{0.44\linewidth}
    \centering
    \includegraphics[width=\linewidth]{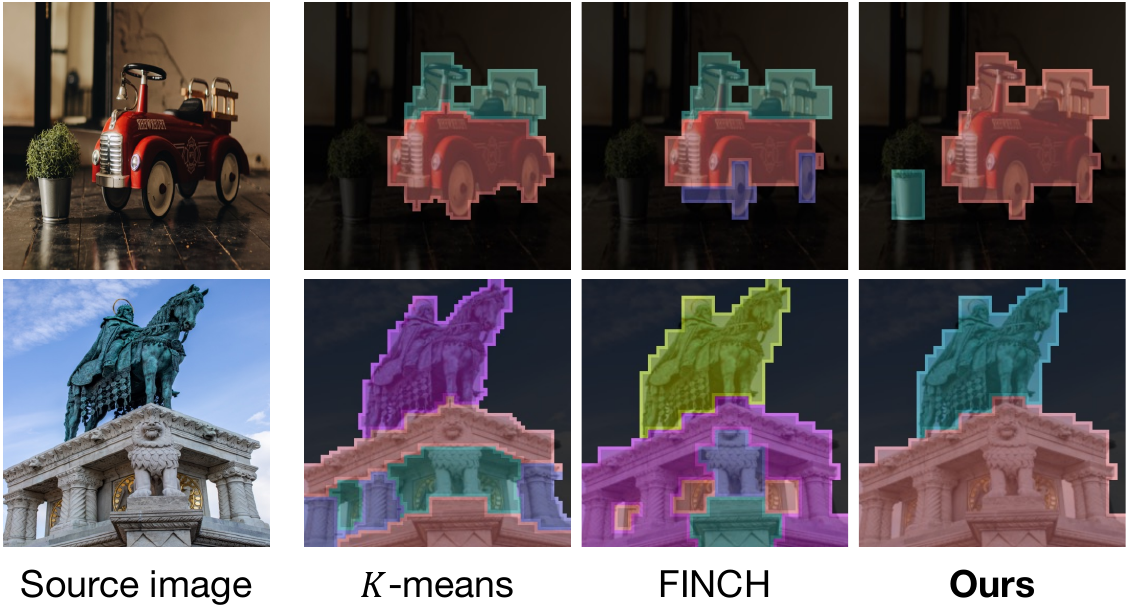}
    \caption{\textbf{Comparison on self-attention clustering.} Each located concept is enclosed within a distinct colored region.}
    \label{fig:ab-clustering}
\end{minipage}
\vspace{-10pt}
\end{figure}

\rvs{We conduct a quantitative ablation study on the training components in~\cref{tab:ab-all}. 
\paragraph{Effectiveness of split-and-merge strategy (SnM)} By comparing Rows (0) and (1), we validate the benefit of the split-and-merge strategy to initializer-absent training. The split-and-merge strategy effectively improves identity similarity and classification accuracy while slightly sacrificing compositional similarity due to its strong focus on a single concept. In~\cref{fig:ab-split}, we present the generated images at different training steps, which reveals how SnM rectifies the training direction. The results illustrate that SnM effectively expands the concept space, allowing learnable tokens to explore a wider range of concepts, ultimately resulting in a more faithful concept indicator. 
\paragraph{Effectiveness of regularization} By comparing Rows (0) and (2) in~\cref{tab:ab-all}, we observe that regularizing the attention map can enhance the generation performance of individual concepts. Row (3) is our full method which further improves the performance regarding all metrics compared to incorporating each component in Rows (1) and (2). The thorough ablation study indicates the effectiveness of each training component in \modelname.
}

\begin{table}[t]
  \centering
  \begin{minipage}[t]{0.47\linewidth}\centering
  \caption{\rvs{\textbf{Ablation study.} Concept-wise optimization, split-and-merge strategy, and regularization are respectively abbreviated as CwO, SnM, and Reg. The results are evaluated on $D_2$ using DINO.}\vspace{-5pt}}
  \setlength{\tabcolsep}{4pt}
  \scalebox{0.66}{
  \begin{tabular}{lccccccc}
    \toprule
    & CwO & SnM & Reg & \textit{SIM}$^\text{I}$ & \textit{SIM}$^\text{C}$ & \textit{ACC}$^1$ & \textit{ACC}$^3$ \\
    \midrule
    (0) & \cmark &  &  & 0.344 & \textbf{0.549} & 0.625 & 0.776\\
    (1) & \cmark & \cmark &  & 0.362 & 0.519 & 0.750 & 0.901\\
    (2) & \cmark &  & \cmark & 0.364 & 0.490 & 0.724 & 0.895 \\
    (3) & \cmark & \cmark & \cmark & \textbf{0.371} & 0.535 & \textbf{0.803} & \textbf{0.934}\\
    \bottomrule
\end{tabular}}
\label{tab:ab-all}
\end{minipage}\ \ 
\begin{minipage}[t]{0.51\linewidth}\centering
  \caption{\rvs{\textbf{Comparison of concept localization.} ``\#clust'' predefines the cluster number for $k$-kmeans and FINCH. The best value is highlighted in \textbf{bold}, while the second-best value is \underline{underlined}.}\vspace{-5pt}}
  \setlength{\tabcolsep}{2.5pt}
  \scalebox{0.62}{
  \begin{tabular}{lcccccccccc}
    \toprule
    Method & \multicolumn{4}{c}{$k$-means} & \multicolumn{4}{c}{FINCH} & Ours & \textcolor{Gray}{MC~\cite{wang2023cut}} \\
    \cmidrule(lr){2-5} \cmidrule(lr){6-9} \cmidrule(lr){10-10} \cmidrule(lr){11-11}
    \#clust & 4 & 5 & 6 & 7 & 4 & 5 & 6 & 7 & auto & \textcolor{Gray}{given} \\
    \midrule
    IoU & 52.8 & 51.4 & 48.7 & 47.6 & 50.7 & \underline{54.8} & 54.6 & 51.5 & \textbf{57.3} & \textcolor{Gray}{58.1} \\
    Recall & 70.3 & 90.1 & 95.5 & 95.7 & 83.7 & 93.4 & \underline{96.9} & \textbf{97.9} &  89.1 & \textcolor{Gray}{97.3}\\
    Precision & 92.7 & 88.0 & 81.5 & 75.4 & \textbf{98.7} & 90.7 & 85.8 & 77.6 & \underline{93.7} & \textcolor{Gray}{77.0} \\
    \bottomrule
\end{tabular}}
\label{tab:ab-clustering}
\end{minipage}
\end{table}

\vspace{-5pt}
\subsection{Concept Localization Analysis}
\paragraph{Self-attention clustering}
To validate the significance of our three-phase method for concept localization, we compare it with $k$-means and our base method FINCH~\cite{sarfraz2019efficient}. Since $k$-means requires a predefined cluster number and FINCH requires a stopping point, we set a proper cluster number of 7 for them. After clustering, we apply the proposed filtering method for fair comparison. We compare the concept localization results using $k$-means, FINCH, and our method in~\cref{fig:ab-clustering}. We can observe that using $k$-means or FINCH will miss some concepts (the 1st row) or split a single concept (the 2nd row). In contrast, our method effectively locates complete and intact concepts, automatically determining a reasonable concept number.
\rvs{
\paragraph{Concept localization benchmark} To quantitatively evaluate the concept localization performance, we establish a benchmark by (1) building a test dataset of multi-concept images along with ground-truth concept masks, and (2) devising tailored metrics for concept localization.
The test dataset is sourced from CLEVR~\cite{johnson2017clevr}, a synthetic dataset featuring clean backgrounds and clear, distinct objects. In this dataset, each object explicitly represents a concept, thereby eliminating potential discrepancies in human-defined concepts in natural images. By comparing the predicted masks with the ground-truth masks in the test set using the Hungarian algorithm~\cite{kuhn1955hungarian}, we can evaluate three metrics: (1) Intersection over Union (IoU) that assesses segmentation accuracy, (2) Recall that evaluates the proportion of true concepts the model can discover, and (3) Precision that evaluates the correctness of the discovered concepts. We provide additional details of this evaluation benchmark for concept localization in Appendix~\textcolor{red}{B}.}
\rvs{
\paragraph{Quantitative evaluation}
Based on the established benchmark, we evaluate our method compared to $k$-means and FINCH with various predefined cluster numbers in~\cref{tab:ab-clustering}. \crhsz{We also report the results of a training-free segmentation method MaskCut (MC) introduced in CutLER~\cite{wang2023cut} for reference, which performs comparably to our method.} When using $k$-means and FINCH, the predefined cluster number can significantly impact performance, and no fixed number can consistently achieve desired performance across all metrics. In contrast, our method performs well in terms of IoU and Precision, with a slight trade-off in Recall. One possible reason is that, unlike specifying the cluster number, our model automatically determines it. Therefore, there may be cases where two concepts are merged into one, resulting in one concept being unable to be matched to the ground truth, potentially reducing recall. Nevertheless, as the only method capable of automatically determining the number of concepts, our method achieves the best overall performance compared to all other clustering techniques.
}
\vspace{-7pt}
\subsection{Unsupervised \vs Supervised}
Although \modelname is an unsupervised model, it would be intriguing to compare \modelname to some supervised methods. Motivated by this, we experiment by providing initial words and ground-truth object masks (obtained by SAM~\cite{kirillov2023segment}) for the supervised method Break-A-Scene~\cite{avrahami2023break}. We compare our method, BaS$^\dag$, and three supervised methods by adding different supervision to Break-A-Scene, as shown in~\cref{fig:ab-sup}. We can observe that adding initial words guides the generation towards the specified category, while adding ground-truth object masks enhances the preservation of texture details. However, even with these two settings, the generated results of Break-A-Scene still fall short compared to our unsupervised model. Despite being trained in a completely unsupervised manner, our model performs on par with the fully supervised setting, where both types of supervision are used. This result further highlights the effectiveness of our model in addressing the concept extraction problem.

\vspace{-5pt}
\subsection{Text-prompted Generation}
\begin{figure}[t]
\begin{minipage}[b]{0.49\linewidth}
    \centering
    \includegraphics[width=\linewidth]{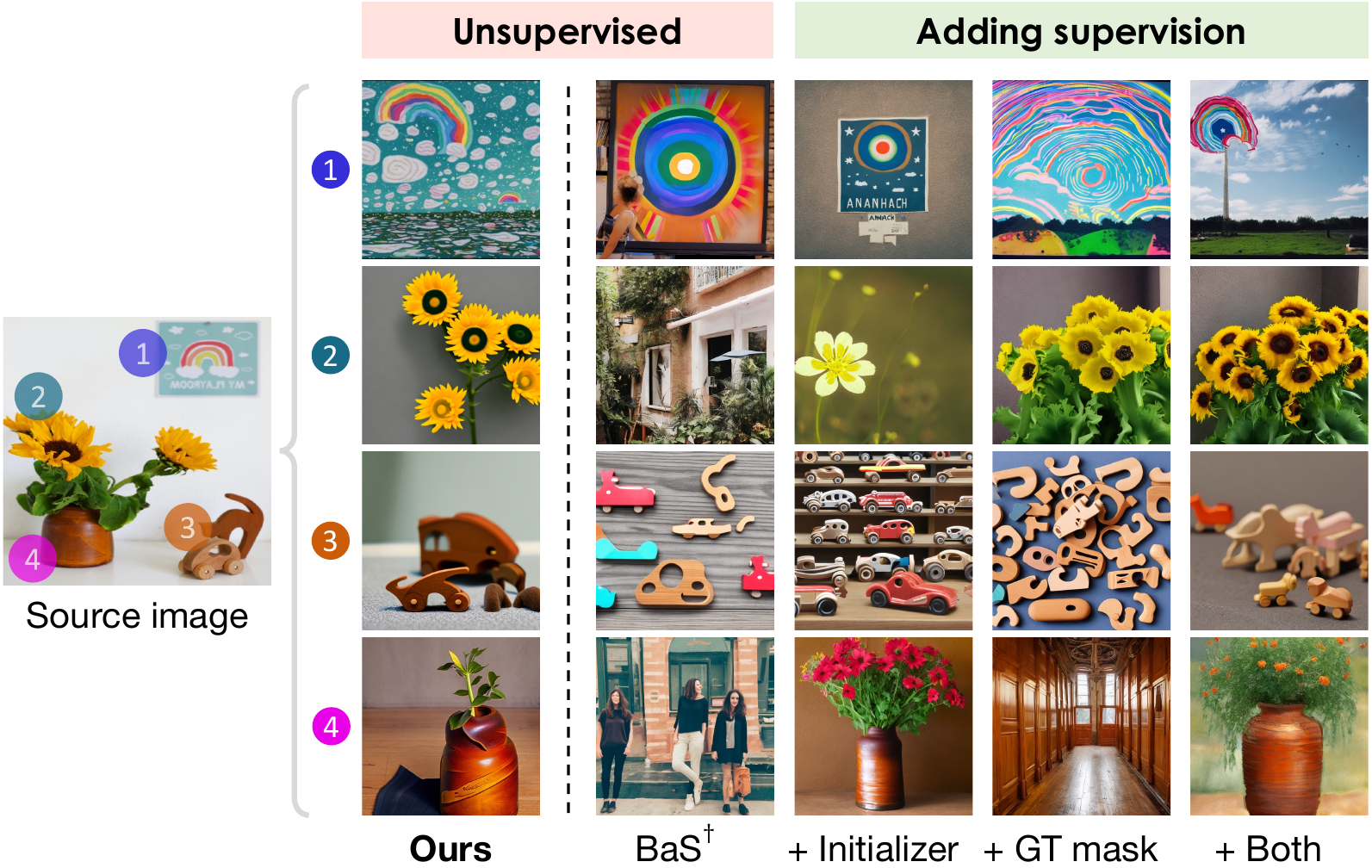}
    \caption{\textbf{Comparison with supervised methods.} We compare \modelname and BaS$^\dag$ to the supervised methods of Break-A-Scene~\cite{avrahami2023break} with added initializers, ground truth masks, and both of them.}
    \label{fig:ab-sup}
\end{minipage}\ \ \ \ 
\begin{minipage}[b]{0.49\linewidth}
    \centering
    \includegraphics[width=\linewidth]{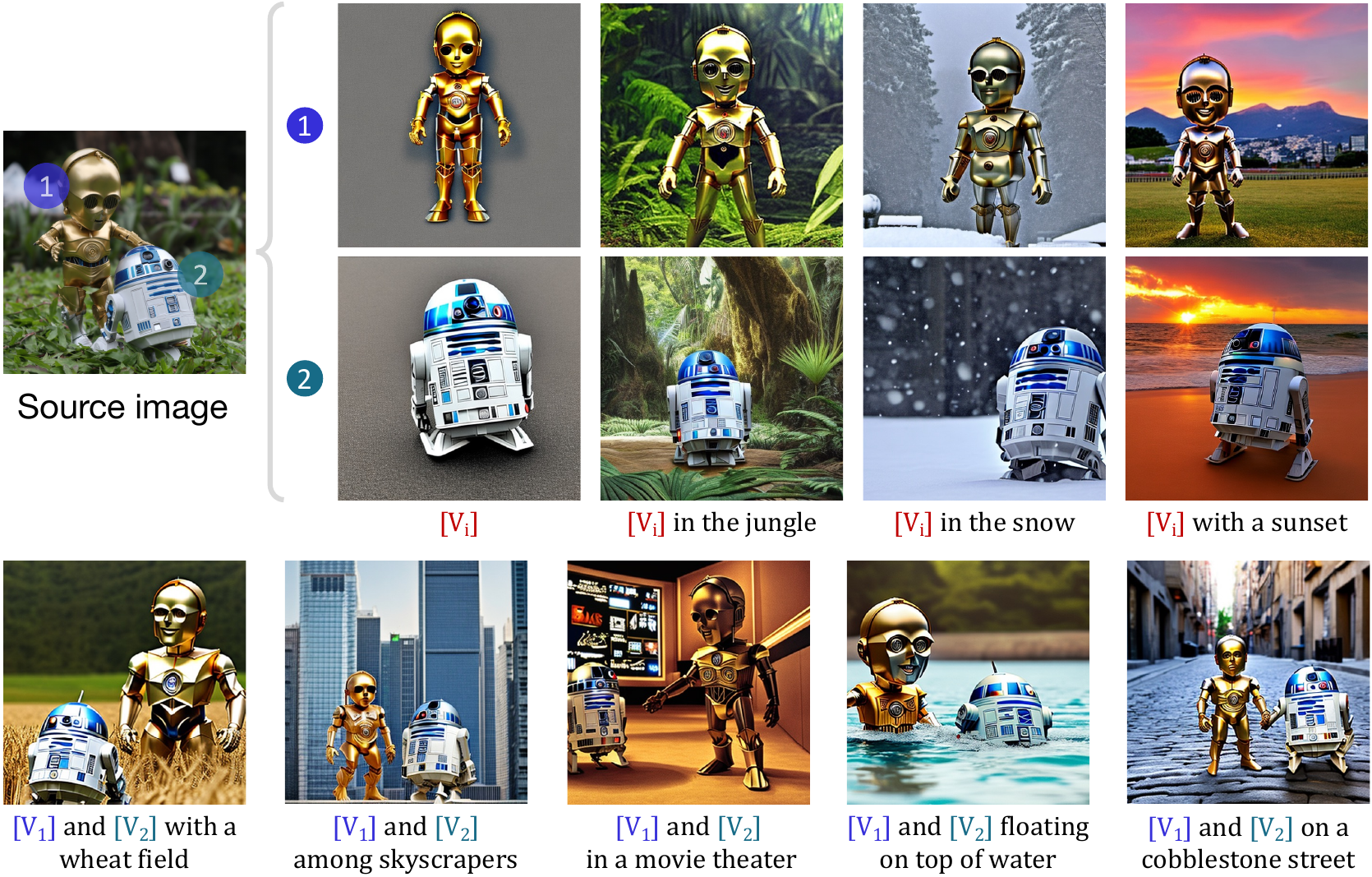}
    \caption{\textbf{Text-prompted generation.} We show the generation results prompted by various text contexts using a single conceptual token (top) and multiple conceptual tokens (bottom).}
    \label{fig:app-prompt}
\end{minipage}
\end{figure}

With the extracted generative concepts, we can perform text-prompted generation. In~\cref{fig:app-prompt}, we showcase the results conditioned on various text prompts using both individual concepts and compositional concepts. The results demonstrate that the learned conceptual tokens can generate images with high text fidelity, aligning faithfully with the text prompt. Furthermore, the images generated with the conceptual tokens also preserve consistent concept identity with the source concepts in both individual and compositional generation. Please refer to Appendix~\textcolor{red}{F} for additional photorealistic results of text-prompted generation.

\section{Conclusion}
\label{sec:conclusion}
In this paper, we introduce Unsupervised Concept Extraction (UCE) that aims to leverage diffusion models to learn individual concepts from a single image in an unsupervised manner. 
We present \modelname to tackle the UCE problem by harnessing the capabilities of pretrained diffusion models to locate concepts and learn their corresponding conceptual tokens. 
Moreover, we establish an evaluation protocol for the UCE problem.
Extensive experiments highlight \modelname as a promising solution to the UCE task.

\par\vfill\par

\clearpage  

\paragraph{Acknowledgement} \crhsz{This work is partially supported by the Hong Kong Research Grants Council - General Research Fund (Grant No.: 17211024).}

%
%
\bibliographystyle{splncs04}
\bibliography{main}

\clearpage
\setcounter{page}{1}
\setcounter{section}{0}

\renewcommand\thesection{\Alph{section}}

\section{More Details on Implementation}
\paragraph{Attention aggregation}
The self-attention maps in different layers have varying resolutions. To aggregate them into a single map for further processing, we follow the approach used in~\cite{tian2023diffuse}. Each self-attention map $\mathbf{A}_l[I,J,:,:]$ represents the correlation between the location $(I,J)$ and all spatial locations. As a result, the last two dimensions of the self-attention maps have spatial consistency, and we interpolate them to ensure uniformity. On the other hand, the first two dimensions of the self-attention maps indicate the locations to which attention maps refer. Therefore, we duplicate these dimensions. By interpolation and duplication, we align the self-attention maps in all layers to a common latent resolution (\ie, 64$\tiny{\times}$64). Finally, we compute the average of all attention maps to obtain the aggregated map. This aggregation step allows us to create a unified attention map combining all maps from different layers.

\paragraph{Conceptual token learning} 
We utilize the split-and-merge strategy in training conceptual tokens. In the training after splitting, we not only sample prompts of individual tokens but also sample a compositional prompt ``a photo of $\mathtt{[V_1]}$ and $\mathtt{[V_2]}$ ... $\mathtt{[V_N]}$'' for training. This approach enhances the compositionality of the learnable tokens. However, unlike~\cite{avrahami2023break}, we refrain from using all possible compositions and instead only use the full composition. This decision is made to ensure that the proportion of using single tokens remains high during training. This, in turn, facilitates effective learning of each individual conceptual token.

\paragraph{Earth mover's distance (EMD)}
We penalize attention alignment using the location-aware EMD.
The EMD is formulated as an optimal transportation problem. Suppose we have supplies of $n_s$ sources $\mathcal{S}=\{s_i\}_{i=1}^{n_s}$ and demands of $n_d$ destinations $\mathcal{D}=\{d_j\}_{j=1}^{n_d}$. Given the moving cost from the $i$-th source to the $j$-th destination $c_{ij}$, an optimal transportation problem aims to find the minimal-cost flow $f_{ij}$ from sources to destinations:
\begin{align}
    \underset{f_{ij}}{\text{minimize}} \quad &\sum\nolimits_{i=1}^{n_s}\sum\nolimits_{j=1}^{n_d} c_{ij} f_{ij} \\
    \text{subject to} \quad &f_{ij} \geqslant 0,\ i=1, ..., n_s,\ j= 1, ..., n_d \\
    &\sum\nolimits_{j=1}^{n_d} f_{ij} = s_i, \ i=1, ..., n_s \\
    &\sum\nolimits_{i=1}^{n_s} f_{ij} = d_j, \ j=1, ..., n_d
\end{align}
where the optimal flow $\tilde{f}_{ij}$ is computed by the moving cost $c_{ij}$, the supplies $s_i$, and the demands $d_j$. The EMD can be further formulated as $(1-c_{ij})\tilde{f}_{ij}$. In our problem, the cross-attention map $\mathbf{c}_\mathtt{[V_i]^j}$ represents the supply, while the target mean attention $\mathbf{f}_i$ represents the demand.
The moving cost is calculated as the Euclidean distance between spatial locations.
Unlike the MSE, the EMD considers differences not only between elements at the same location but also between elements at different locations. This means that the EMD takes into account both spatial alignment and the magnitude of differences, providing a more comprehensive measure of dissimilarity.

\section{Concept Localization Benchmark}
\rvs{In the main paper, we present a novel benchmark to evaluate concept localization performance. This benchmark effectively assesses our model's concept localization capability in two aspects: (1) concept discovery accuracy and (2) concept segmentation efficacy. Here, we offer further details on the benchmark, including the dataset and evaluation metrics.}
\begin{figure}[t]
    \centering
    \includegraphics[width=\linewidth]{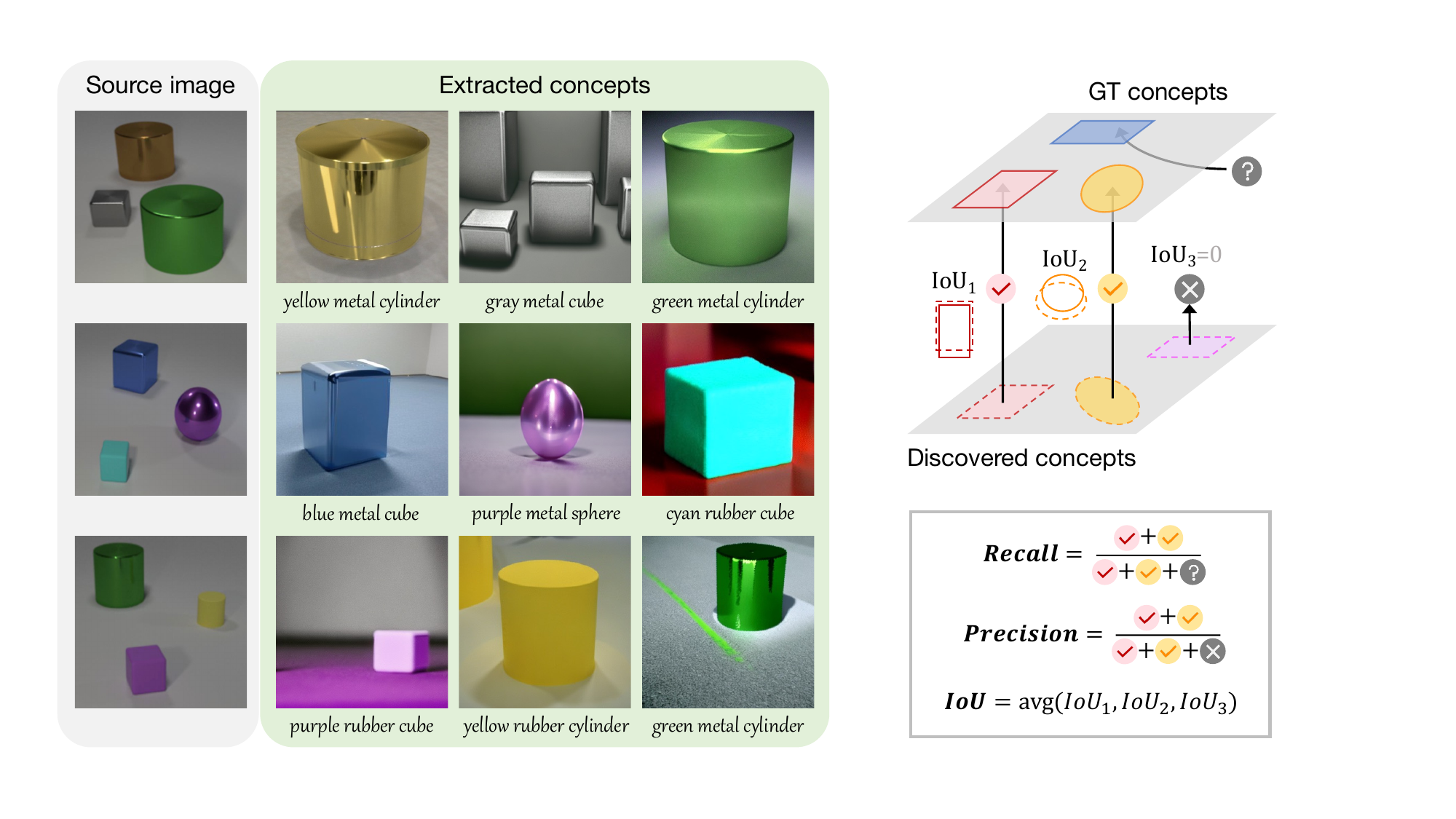}
    \caption{\rvs{\textbf{Concept localization benchmark.}
    \textbf{Left}: We present some examples of the source images in the dataset and their corresponding concept-wise images generated by \modelname. We provide the visual properties of each distinct concept underneath the generated image.
    \textbf{Right}: We visualize the matching process between ground-truth (GT) concepts and discovered concepts. Here, \markcheckone and \markchecktwo denote the true positive matches, \marktimes denotes the wrongly discovered concept, and \markquestion denotes the missed true concept.}}
    \label{fig:supp-clevr}
\end{figure}

\paragraph{Dataset curation}
\rvs{To assess concept localization, the benchmark dataset must meet two criteria: (1) clear and distinct concept definition, and (2) accurate ground-truth concept masks. Natural images lack these characteristics. Instead, we source our images from CLEVR~\cite{johnson2017clevr}, a dataset known for its well-defined objects, diverse in colors, materials, and shapes, set against a uniform grey background. We collect a total of 25 images, each containing 3 to 5 concepts, along with their corresponding ground-truth segmentation masks. We show image samples from the dataset alongside the generated images of our extracted concepts in~\cref{fig:supp-clevr} (left).}

\paragraph{Evaluation metrics}
We further introduce evaluation metrics tailored for concept localization. The process of concept localization incorporates two parts: (1)~concept discovery and (2)~concept segmentation. For these two parts, we devise three metrics: \textbf{recall} and \textbf{precision} to assess concept discovery, and average intersection over union (\textbf{IoU}) to assess concept segmentation.
Specifically, let $\mathcal{P}=\{\mathbf{m}_i\}_{i=1}^{N}$ denote the set of the $N$ concept segments discovered by the model, and let $\mathcal{Q}=\{\boldsymbol{\mu}_j\}_{j=1}^{M}$ denote the set of the $M$ ground-truth concept segments. To match the discovered concepts and the ground-truth concepts, we aim to maximize their average inter-instance IoU, which is given by
\begin{equation}
\label{eq:avg-iou}
    \text{Avg. IoU} = \max_{\lambda_1\in\Lambda(M, M^\prime)\atop
 \lambda_2\in\Lambda(N, M^\prime)}\frac{1}{N}\sum_{i=1}^{M^\prime} \text{IoU}\left(\boldsymbol{\mu}_{\lambda_1(i)}, \textbf{m}_{\lambda_2(i)}\right)
\end{equation}
where $M^\prime = \min(M,N)$ and $\Lambda(M,M^\prime)$ denotes the set of all $M^\prime$-permutations of integers ranging from 1 to $M$. Similarly, $\Lambda(N,M^\prime)$ represents the set of all $M^\prime$-permutations of integers ranging from 1 to $N$. We use Hungarian optimal assignment algorithm~\cite{kuhn1955hungarian} to solve the maximization problem of \cref{eq:avg-iou} over the set of permutations. The maximum value of the average IoU reflects the overall segmentation proficiency of our localized concepts. The permutations $\lambda_1$ and $\lambda_2$ together give the matching correspondence between ground-truth concepts and discovered concepts. We further examine the count of \emph{true positive} concept matches that yield non-zero IoUs by
\begin{equation}
    R = \sum_{i=1}^{M^\prime}\mathbbm{1}\left\{\text{IoU}\left(\boldsymbol{\mu}_{\lambda_1(i)}, \textbf{m}_{\lambda_2(i)}\right) \neq 0\right\}
\end{equation}
Therefore, we can compute recall and precision by
\begin{equation}
    \text{recall} = \frac{R}{M} \quad \text{precision} = \frac{R}{N}
\end{equation}
With recall and precision, we can evaluate concept discovery performance based on whether there are missed true concepts or wrongly discovered concepts. The computation of all three evaluation metrics is visualized in~\cref{fig:supp-clevr} (right).

\section{User Study}
\begin{figure}[t]
    \centering
    \includegraphics[width=\linewidth]{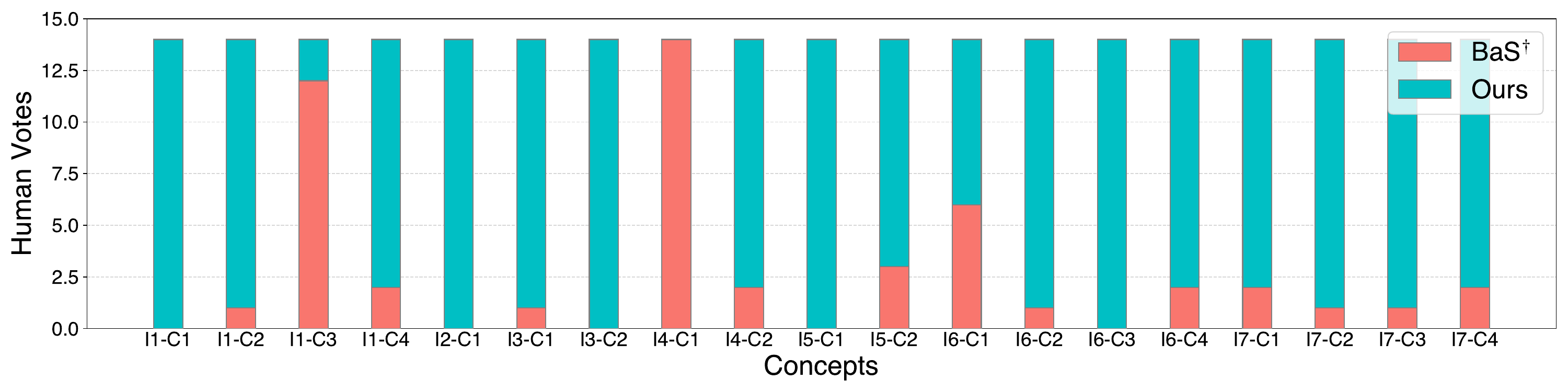}
    \caption{\rvs{\textbf{User study statistics.} We report the human votes of a total of 19 concepts extracted from 7 source images, comparing BaS$^\dag$ and our method. On the horizontal axis, I$\mathtt{n}$-C$\mathtt{m}$ denotes the $\mathtt{m}$-th concept in the $\mathtt{n}$-th image, while the vertical axis displays the number of human votes for each method, represented by different colors.}}
    \label{fig:supp-user-study}
\end{figure}

To ensure the assessment of generation quality aligns with human preference, we conduct a user study comparing the generated results from BaS$^\dag$ and \modelname.  
We asked 14 users to vote between our method and BaS$^\dag$ by viewing the generated images of 19 concepts from 7 images in $D_2$. For each concept, we presented the users with 8 images, randomly generated by \modelname and BaS$^\dag$ respectively, along with the masked image of the source concept. They were then asked to indicate which model produced images that better resembled the source concept.
Finally, we collected a total of 266 user votes, representing human preference. Among all the votes, 18.8\% of the votes favored BaS$^\dag$ while 81.2\% preferred our model. Detailed statistics of the votes for each concept are present in~\cref{fig:supp-user-study}. The user study further indicates that \modelname outperforms BaS$^\dag$ in generating concept images that align with human judgment.

\section{Additional Ablation Studies} 
\subsection{Self-attention Clustering}
\begin{figure}[t]
    \centering
    \includegraphics[width=0.8\linewidth]{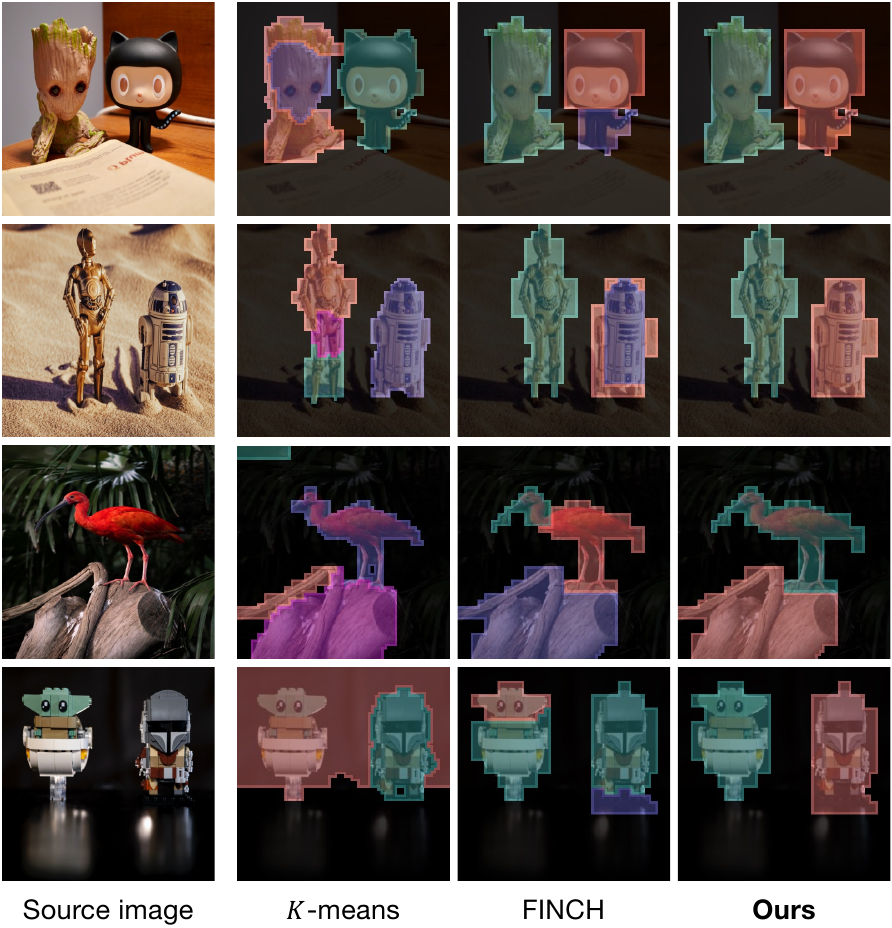}
    \caption{Concept localization results using different methods for self-attention clustering.}
    \label{fig:supp-ab-clustering}
\end{figure}

In~\cref{fig:supp-ab-clustering}, we present additional self-attention clustering results, comparing our approach with $k$-means and FINCH~\cite{sarfraz2019efficient}. 
Upon observation, we note that $k$-means and FINCH may separate a single concept (as seen in the 1st, 2nd, and 3rd rows) or include background regions (as seen in the 4th row) within their clusters. In contrast, our approach consistently demonstrates high accuracy in locating each concept, ensuring precise concept localization within the image.

\subsection{Split-and-merge Strategy} 
\begin{figure}[t]
    \centering
    \includegraphics[width=0.99\linewidth]{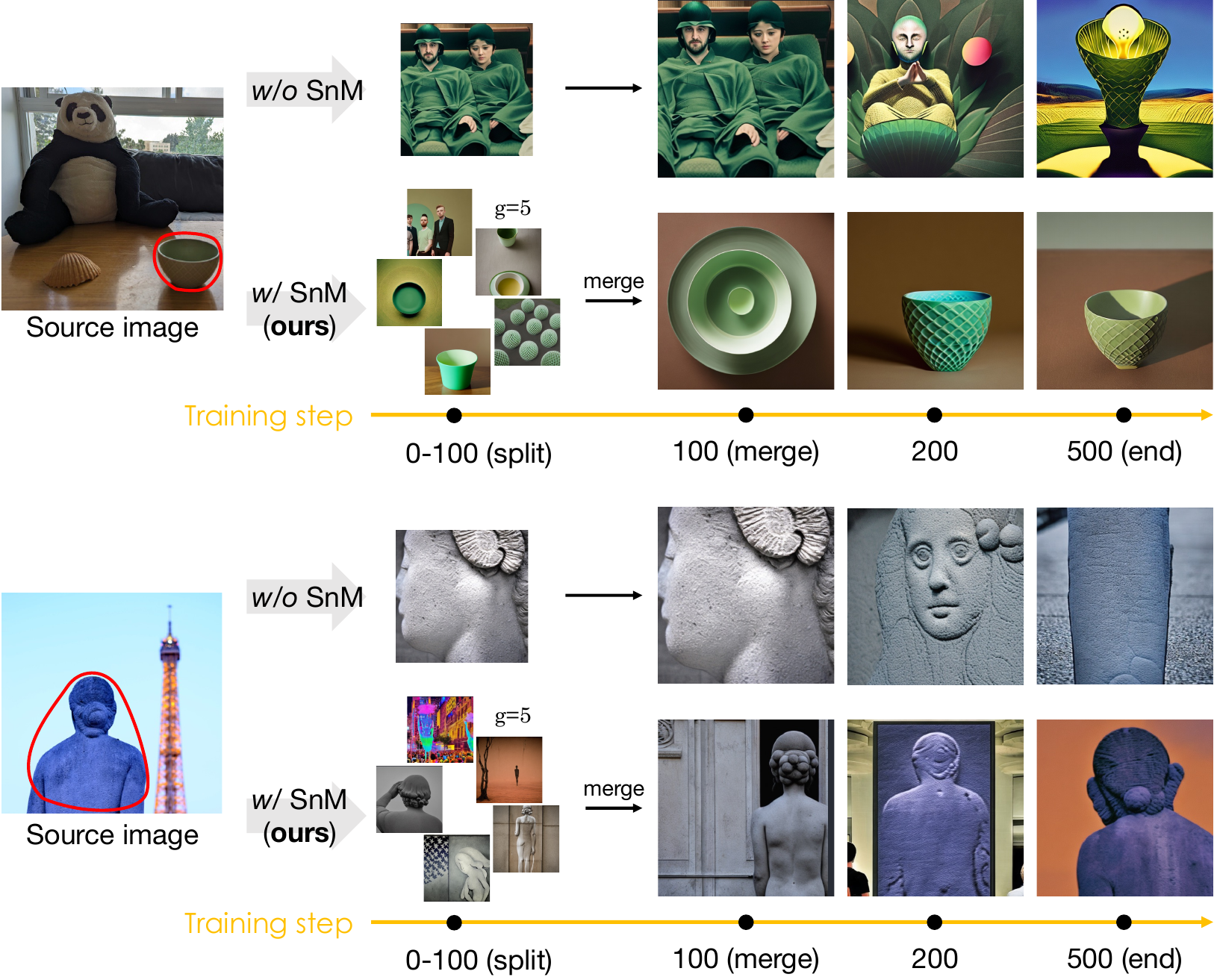}
    \caption{\textbf{Generation results of split-and-merge ablation.}}
    \label{fig:supp-ab-split}
\end{figure}

\paragraph{Visualization} In~\cref{fig:supp-ab-split}, we provide two additional examples illustrating how the split-and-merge strategy rectifies the token learning process. The split-and-merge strategy expands the search space for concepts during the splitting phase, allowing the merged token to exhibit concept characteristics that closely align with the source concept. This improves the ability to learn and represent unseen concepts without initialization, ultimately enhancing image generation quality.

\begin{figure}[t]
    \centering
    \includegraphics[width=\linewidth]{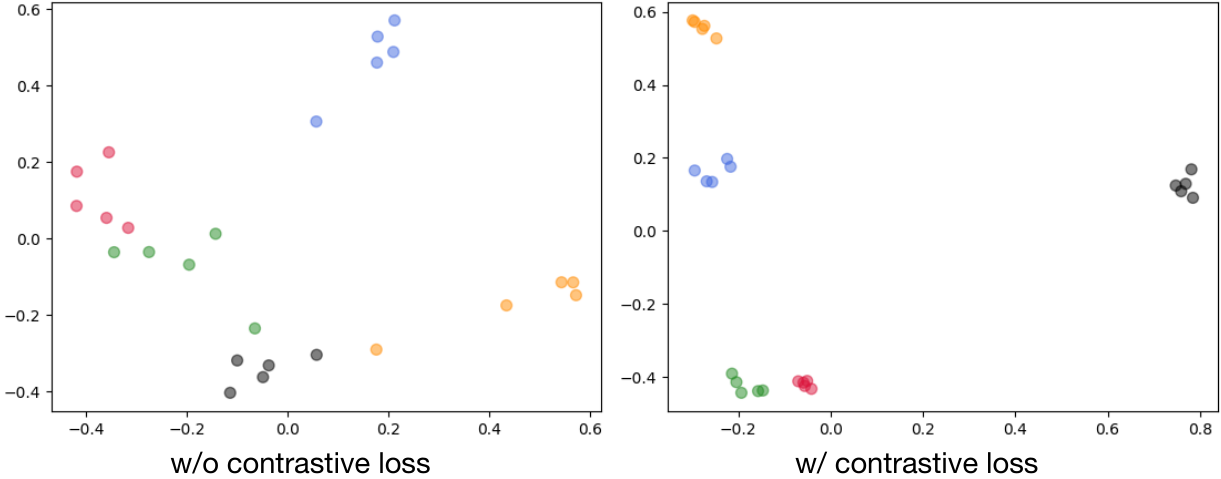}
    \caption{\textbf{Visualization of token embedding vectors.} We randomly initialize 5 embedding vectors marked with the same color to represent each of the 5 concepts, resulting in a total of 25 embedding vectors. After the splitting phase, specifically at step 100, we use PCA to visualize the learned embedding vectors. We compare the results with and without the contrastive loss.}
    \label{fig:supp-pca}
\end{figure}

\paragraph{Effect of contrastive loss} We further investigate the impact of the contrastive loss during the splitting phase. The contrastive loss encourages split tokens representing the same concept to be closer together, promoting better concept agreement across all tokens.
In~\cref{fig:supp-pca}, we employ PCA to visualize the embedding vectors with and without the contrastive loss. When the contrastive loss is used, the embedding vectors representing the same concept exhibit a more compact distribution, aligning with our goal of enhancing concept representation.
We also report the quantitative comparison in~\cref{tab:supp-ab-con}. The use of contrastive loss can enhance the performance on all metrics, especially on classification accuracy.
\begin{table}[t]
\caption{Contrastive loss ablation on $D_1$ using DINO.}
  \label{tab:supp-ab-con}
  \centering
\setlength{\tabcolsep}{10pt}
  \begin{tabular}{lcccc}
    \toprule
    Method & \textit{SIM}$^\text{I}$ & \textit{SIM}$^\text{C}$ & \textit{ACC}$^1$ & \textit{ACC}$^3$ \\
    \midrule
    w/o contrast. & 0.316 & 0.567 & 0.320 & 0.456 \\
    w/ contrast. (ours) & \textbf{0.319} & \textbf{0.568} & \textbf{0.324} & \textbf{0.470} \\
    \bottomrule
  \end{tabular}
\end{table}

\paragraph{Effect of token merging} \rvs{To validate the necessity of merging multiple tokens midway through optimization, we additionally evaluate a variant that optimizes the multiple randomly initialized tokens separately and merges them at the end. We compare the results in~\cref{tab:separate-variant}. Our method significantly outperforms the variant, especially in compositional similarity and classification accuracy.}
\begin{table}[t]
\begin{minipage}[t]{0.49\linewidth}\centering
\caption{\rvs{We present the results of comparison with the variant (\texttt{Var}) that optimizes multiple split tokens separately and merges them at the end. Experiments are conducted on $D_2$ using DINO.}}
  \label{tab:separate-variant}
  \centering
  \setlength{\tabcolsep}{5pt}
  \scalebox{1.0}{
  \begin{tabular}{lcccc}
    \toprule
    & \textit{SIM}$^\text{I}$ & \textit{SIM}$^\text{C}$ & \textit{ACC}$^1$ & \textit{ACC}$^3$ \\
    \midrule
    \texttt{Var} & 0.369 & 0.400 & 0.730 & 0.868 \\
    Ours & \textbf{0.371} & \textbf{0.535} & \textbf{0.803} & \textbf{0.934}\\
    \bottomrule
\end{tabular}}
\end{minipage}\ \ 
\begin{minipage}[t]{0.49\linewidth}\centering
\caption{\rvs{Results vary with different numbers of split tokens. Using only a single token ($g$=1) yields \textcolor{Red}{poor} performance. Experiments are conducted on $D_2$ using DINO.}}
  \label{tab:num-tok}
  \centering
  \setlength{\tabcolsep}{3pt}
  \scalebox{0.9}{
  \begin{tabular}{lcccc}
    \toprule
    \#tokens & \textit{SIM}$^\text{I}$ & \textit{SIM}$^\text{C}$ & \textit{ACC}$^1$ & \textit{ACC}$^3$ \\
    \midrule
     $g$=1 & \textcolor{Red}{0.319} & \textcolor{Red}{0.486} & \textcolor{Red}{0.651} & \textcolor{Red}{0.789} \\
     $g$=3 & \textbf{0.381} & \textbf{0.535} & 0.750 & 0.914 \\
     $g$=5 (ours) & 0.371 & \textbf{0.535} & \textbf{0.803} & \textbf{0.934} \\
     $g$=7 & 0.367 & 0.525 & 0.743 & 0.882 \\
    \bottomrule
\end{tabular}}
\end{minipage}
\end{table}

\paragraph{Effect of the number of split tokens} \rvs{To observe the impact of the number of split tokens, \ie, the value of $g$, on the training process, we evaluate the performance of our model using different numbers of split tokens. The results are reported in~\cref{tab:num-tok}. From the table, we can make the following observations: 
(1) Using only a single token ($g$=1), \ie, excluding the split-and-merge strategy, yields poor performance, again demonstrating the significance of the split-and-merge strategy; 
(2) $g$=3 performs comparably to our setting ($g$=5), with a slight improvement in identity similarity and a moderate drop in classification accuracy; 
(3) A larger number (\eg, $g$=7) may decrease the performance across all metrics to some extent.
In the main paper, we set $g$=5 to balance all metrics. This experiment underscores the significance of our split-and-merge strategy for robust token initialization, which can greatly impact performance.
}

\subsection{Attention Alignment}
\begin{figure}[t]
    \centering
    \includegraphics[width=0.95\linewidth]{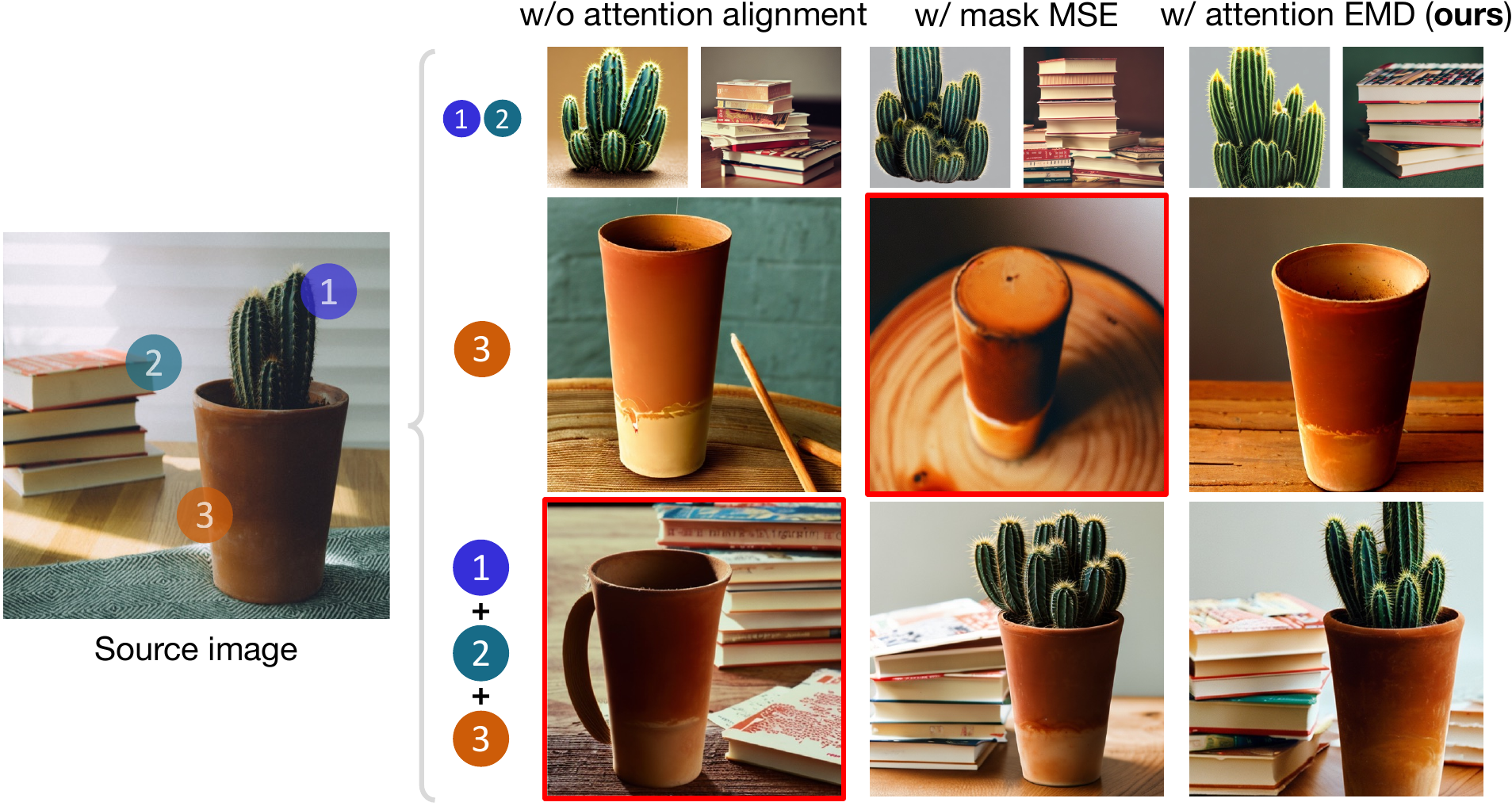}
    \caption{Individual and compositional generation results of attention comparison.}
    \label{fig:supp-ab-alignment}
\end{figure}

\rvs{We compare training with EMD attention alignment (ours) to training with MSE attention alignment and training without attention alignment, as present in~\cref{fig:supp-ab-alignment}.} From the observation, we can see that when attention alignment is not used, some concepts in the compositional generated image may be missed. On the other hand, using mask MSE can lead to unsatisfactory single-concept generations. In contrast, our EMD attention alignment strikes a balance between these two extremes and performs well in both aspects. It ensures that the model captures and represents all the desired concepts in the compositional image while also generating high-quality single-concept images. Our attention alignment approach achieves a favorable trade-off.

\section{Additional Quantitative Analysis} 
\subsection{Unsupervised \vs Supervised} We present quantitative results for unsupervised methods, namely \modelname and BaS$^\dag$, as well as methods augmented with different types of supervision. The comparison results are presented in~\cref{tab:supp-unsup-sup-quant}. \rvs{By comparing (5) to (1)-(3), we can observe that unsupervised \modelname outperforms supervised versions of BaS$^\dag$. We provide supervision of ground-truth SAM masks for our model in (6) which slightly improves \textit{SIM}$^\text{I}$, \textit{SIM}$^\text{C}$, and \textit{ACC}$^3$ while lowering \textit{ACC}$^1$. This result indicates that our identified masks exhibit performance characteristics largely comparable to SAM masks. We further finetune the fully supervised version of BaS$^\dag$ in (4), which is the \emph{original implementation} of Break-A-Scene~\cite{avrahami2023break}. We also finetune our unsupervised model in (7). The results clearly show that, with finetuning, our unsupervised model significantly outperforms the originally implemented BaS$^\dag$ that is fully supervised by masks and initial words.}
\begin{table}[t]
\caption{\textbf{Unsupervised \vs supervised.} We compare {\sethlcolor{Cyan!10}\hl{unsupervised}} settings, \ie, \modelname and BaS$^\dag$, and {\sethlcolor{Yellow!10}\hl{supervised}} settings by adding different supervision to Bas$^\dag$, including the ground-truth SAM masks ($\tiny{+}$Mask) and the human-annotated initial words ($\tiny{+}$Init.). We additionally report the results of finetuning ($\tiny{+}$FT) the whole diffusion model for reference. \rvs{Row (4) represents the original implementation of Break-A-Scene~\cite{avrahami2023break}.} Experiments are conducted on $D_2$ using DINO.} 
  \label{tab:supp-unsup-sup-quant}
  \centering
  \setlength{\tabcolsep}{5pt}
  \scalebox{0.9}{
  \begin{tabular}{lcccccccc}
    \toprule
     & Method & $\tiny{+}$Mask & $\tiny{+}$Init. & $\tiny{+}$FT & \textit{SIM}$^\text{I}$ & \textit{SIM}$^\text{C}$ & \textit{ACC}$^1$ & \textit{ACC}$^3$ \\
    \midrule
    (1) & \multirow{4}{*}{BaS$^\dag$} & \cellcolor{Cyan!10} & \cellcolor{Cyan!10} & \cellcolor{Cyan!10} & \cellcolor{Cyan!10} 0.231 & \cellcolor{Cyan!10} 0.417 & \cellcolor{Cyan!10} 0.329 & \cellcolor{Cyan!10} 0.559\\
     (2) & & \cellcolor{Yellow!10} \cmark & \cellcolor{Yellow!10} & \cellcolor{Yellow!10} & \cellcolor{Yellow!10} 0.266 & \cellcolor{Yellow!10} 0.430 & \cellcolor{Yellow!10} 0.388 & \cellcolor{Yellow!10} 0.618\\
     (3) & & \cellcolor{Yellow!10} \cmark & \cellcolor{Yellow!10} \cmark & \cellcolor{Yellow!10} & \cellcolor{Yellow!10} 0.316 & \cellcolor{Yellow!10} 0.474 & \cellcolor{Yellow!10} 0.559 & \cellcolor{Yellow!10} 0.704\\
     (4) & & \cellcolor{Yellow!10} \cmark & \cellcolor{Yellow!10} \cmark & \cellcolor{Yellow!10} \cmark & \cellcolor{Yellow!10} 0.411 & \cellcolor{Yellow!10} 0.696 & \cellcolor{Yellow!10} 0.697 & \cellcolor{Yellow!10} 0.737\\
    \arrayrulecolor{gray} \hdashline
    (5) & \multirow{3}{*}{Ours} & \cellcolor{Cyan!10} & \cellcolor{Cyan!10} & \cellcolor{Cyan!10} & \cellcolor{Cyan!10} 0.371 & \cellcolor{Cyan!10} 0.535 & \cellcolor{Cyan!10} 0.803 & \cellcolor{Cyan!10} 0.934 \\
     (6) & & \cellcolor{Yellow!10} \cmark & \cellcolor{Yellow!10} & \cellcolor{Yellow!10} & \cellcolor{Yellow!10} 0.396 & \cellcolor{Yellow!10} 0.564 & \cellcolor{Yellow!10} 0.757 & \cellcolor{Yellow!10} 0.974 \\
     (7) & & \cellcolor{Cyan!10} & \cellcolor{Cyan!10} & \cellcolor{Cyan!10} \cmark & \cellcolor{Cyan!10} \textbf{0.490} & \cellcolor{Cyan!10} \textbf{0.785} & \cellcolor{Cyan!10} \textbf{0.888} & \cellcolor{Cyan!10} \textbf{0.980}\\
    \bottomrule
  \end{tabular}}
\end{table}

\begin{table}[t]
  \caption{\textbf{Text prompt set.} ``\{\}'' represents the conceptual token.}
  \label{tab:supp-text-prompt}
  \centering
  \scalebox{0.9}{
  \begin{tabular}{l}
    \toprule
    Prompts \\
    \midrule
    a photo of \{\} in the jungle \\
    a photo of \{\} in the snow \\
    a photo of \{\} at the beach \\
    a photo of \{\} on top of pink fabric \\
    a photo of \{\} on top of a wooden floor \\
    a photo of \{\} with a city in the background \\
    a photo of \{\} with a mountain in the background \\
    a photo of \{\} floating on top of water \\
    a photo of \{\} with a tree and autumn leaves in the background \\
    a photo of \{\} with the Eiffel Tower in the background \\
    a photo of \{\} on top of the sidewalk in a crowded street \\
    a photo of \{\} with a Japanese modern city street in the background \\
    a photo of \{\} on top of a dirt road \\
    a photo of \{\} among the skyscrapers in New York City \\
    a photo of \{\} in a dream of a distant galaxy \\
    \bottomrule
  \end{tabular}}
\end{table}

\subsection{Text Guidance} We also explore the performance of subject-driven text-to-image generation. To do this, we utilize a set of prompts to generate text-conditioned images with all extracted concepts and their compositions.
We expand the set of prompts used in~\cite{avrahami2023break} from 10 to 15 in~\cref{tab:supp-text-prompt}.
We evaluate the generated images by measuring their CLIP image similarity with the masked source image, as well as their CLIP text similarity with the corresponding text prompt in~\cref{tab:supp-text-prompt} (with the learnable token removed). 

\begin{figure}[t]
    \centering
    \includegraphics[width=0.7\linewidth]{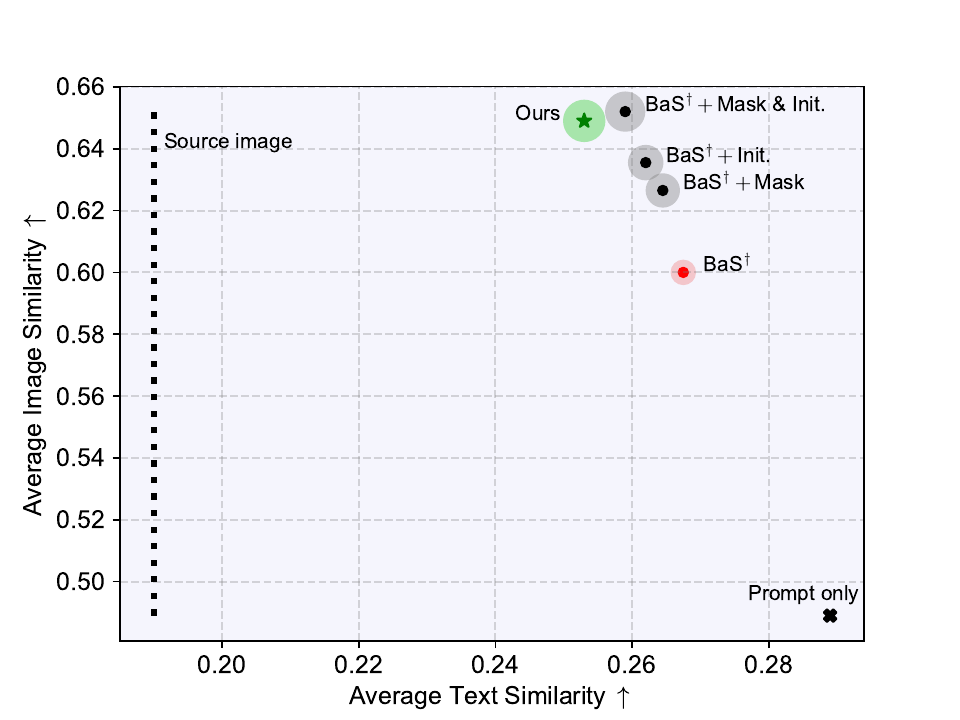}
    \caption{\textbf{Quantitative evaluation of subject-driven text-to-image generation.} The bounding circle size represents the normalized mean value of the main metrics reported in~\cref{tab:supp-unsup-sup-quant}. We also include evaluation results for two additional scenarios: (1) Prompt only: we evaluate all metrics when generating images using only the text prompt; (2) Source image: we evaluate the text similarity between the masked source images and all prompts. }
    \label{fig:supp-curve}
\end{figure}

In~\cref{fig:supp-curve}, we present the average image and text similarities for all compared methods.
We observe that as supervision is gradually added to BaS$^\dag$, the image similarity increases while the text similarity decreases. This is expected since the unsupervised BaS$^\dag$ may struggle to learn certain concepts represented by the learnable tokens, resulting in the text prompt dominating the text embedding space and excessively guiding the image generation process. However, as supervision is introduced, BaS$^\dag$ can better learn the conceptual tokens and prioritize the subject concept in the generated image, thereby reducing the reliance on text information. The lower text similarity indicates the text information becomes less pronounced rather than completely absent.
\modelname also exhibits high image similarity and slightly lower text similarity. Notably, it performs closely to the fully supervised BaS$^\dag$, indicating that \modelname effectively learns reliable generative conceptual tokens for subject-driven image generation.

\subsection{Larger Classifier}
\rvs{In the evaluation of classification accuracy, we can increase the number of prototypes in the classifier by including a large codebook of concepts besides the concepts in the datasets. Specifically, we randomly sample one image per class from ImageNet-1k~\cite{russakovsky2015imagenet}, obtaining 1,000 images in total. We encode them as additional concept prototypes in the classifier. We report the results of classification accuracy by using the larger classifier in~\cref{tab:larger-cls}.
We note that the accuracy values under the larger classifier are considerably lower compared to those under the original classifier for both models. 
Nevertheless, our model consistently demonstrates significant superiority over BaS$^\dag$ across all evaluation criteria, regardless of whether the larger or the original classifier is employed.}
\begin{table}[t]
\captionof{table}{\textbf{Classification accuracy under a larger classifier.} We use {\sethlcolor{Lavender!50}\hl{*}} to denote the results evaluated with the larger classifier that contains 1,000 additional prototypes sampled from ImageNet-1k. Experiments are conducted on $D_2$ evaluated using DINO.}
\setlength{\tabcolsep}{5pt}
  \label{tab:larger-cls}
  \centering
  \scalebox{0.9}{
  \begin{tabular}{lcccc}
    \toprule
    & \textit{ACC}$^1$ & \cellcolor{Lavender!50}{\textit{*ACC}$^1$} & \textit{ACC}$^3$ & \cellcolor{Lavender!50}{\textit{*ACC}$^3$} \\
    \midrule
    BaS$^\dag$ & 0.329 & \cellcolor{Lavender!50}{0.138} & 0.559 & \cellcolor{Lavender!50}{0.203} \\
    Ours & \textbf{0.803} & \cellcolor{Lavender!50}{\textbf{0.395}} & \textbf{0.934} & \cellcolor{Lavender!50}{\textbf{0.546}} \\
    \bottomrule
  \end{tabular}}
\end{table}

\subsection{Initializer Analysis}
\rvs{In our unsupervised setting, initial words for training each concept are inaccessible because the concepts are automatically extracted, and human examination of each concept requires costly labor. We introduce the split-and-merge strategy to address this problem.}

\begin{table}[t]
\captionof{table}{\rvs{\textbf{Comparison of using different types of initializers.} We initialize conceptual tokens using human annotation (Human init.) and model annotation (CLIP init.) by CLIP~\cite{radford2021learning} vocabulary retrieval for BaS$^\dag$ and our model, in comparison to our unsupervised method. Experiments are conducted on $D_2$ using DINO.}}
  \label{tab:various-init}
  \centering
  \setlength{\tabcolsep}{5pt}
  \scalebox{1.0}{
  \begin{tabular}{lcccc}
    \toprule
    & \textit{SIM}$^\text{I}$ & \textit{SIM}$^\text{C}$ & \textit{ACC}$^1$ & \textit{ACC}$^3$ \\
    \midrule
     BaS$^\dag$ w/ Human Init. & 0.312 & 0.506 & 0.586 & 0.730 \\
     BaS$^\dag$ w/ CLIP Init. & 0.304 & 0.470 & 0.507 & 0.658 \\
     Ours w/ Human Init. & \textbf{0.390} & 0.515 & \textbf{0.803} & 0.914 \\
     Ours w/ CLIP Init. & 0.330 & 0.514 & 0.638 & 0.862 \\
     Ours (\emph{unsupervised}) &  0.371 & \textbf{0.535} & \textbf{0.803} & \textbf{0.934} \\
    \bottomrule
\end{tabular}}
\end{table}

\rvs{Despite this, we still aim to explore the performance of models trained with different types of initializers in supervised settings and compare them with our unsupervised approach. We consider two types of initializers: (1) human-annotated initializers and (2) model-annotated initializers. 
For human annotation, we directly assign a suitable word to an extracted concept based on human preference.
For model annotation, we use the vision-language model CLIP~\cite{radford2021learning} to retrieve a word from CLIP text tokenizer's vocabulary. This word is selected based on the highest similarity to the extracted concept, determined by comparing features between each word in the vocabulary set and the masked image part of the concept. We apply these two types of initializers to BaS$^\dag$ and our model, and compare the experimental results in~\cref{tab:various-init}.}

\rvs{We observe the following: (1) Our model consistently outperforms BaS$^\dag$, regardless of the type of initializer; (2) Human annotation generally yields better performance than model annotation for both models; (3) Although our unsupervised approach slightly lags behind our model with the use of human-annotated initializers in terms of identity similarity, it still outperforms all other supervised methods in all metrics. These observations demonstrate the effectiveness of the split-and-merge strategy in resolving the challenge of inaccessible initializers in unsupervised concept extraction.}

\section{Additional Comparison and Our Results}
\begin{figure*}
    \centering
    \includegraphics[width=0.99\linewidth]{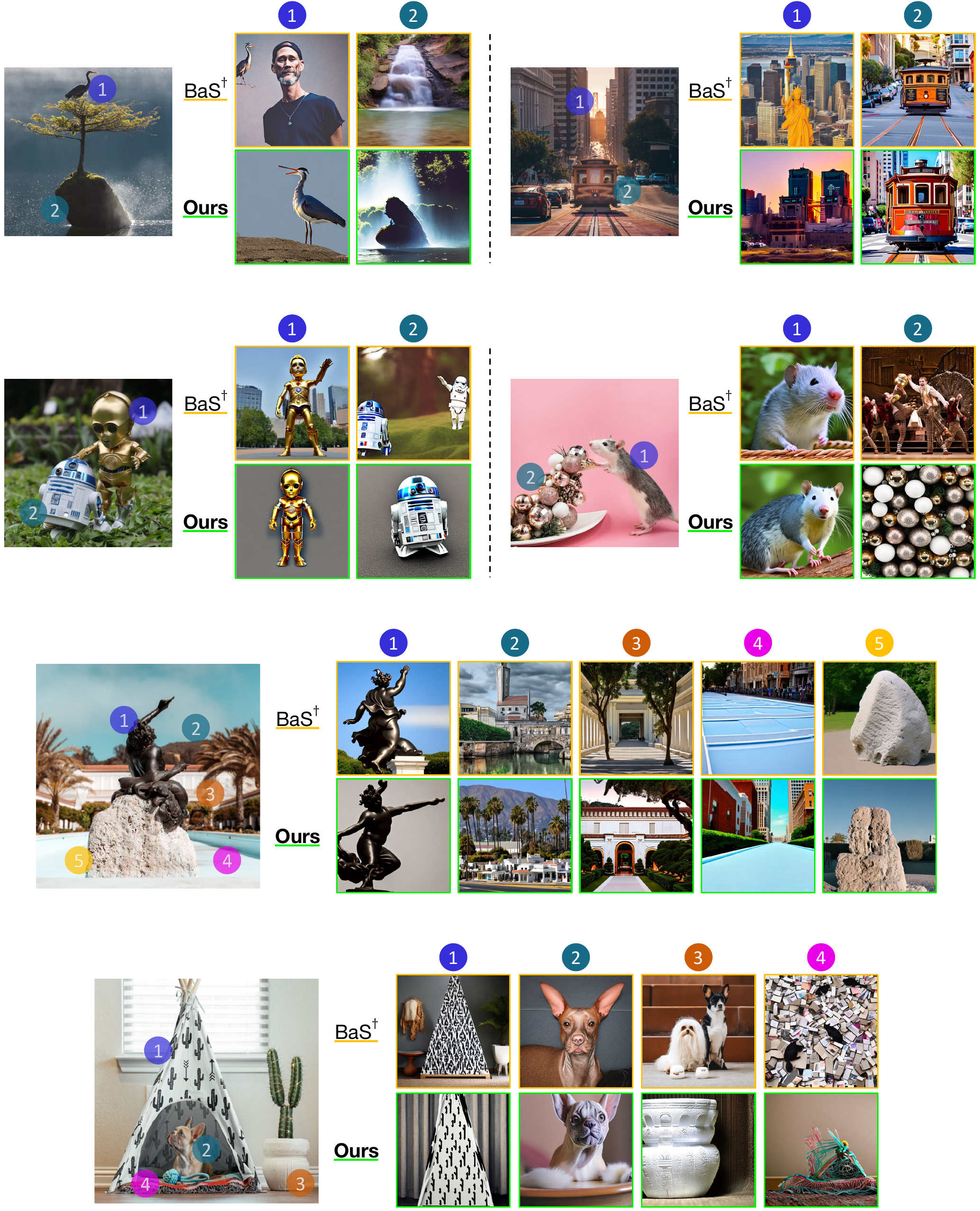}
    \caption{\textbf{Additional results comparing between \modelname and BaS$^\dag$.}}
    \label{fig:supp-compare}
\end{figure*}

\begin{figure*}
    \centering
    \includegraphics[width=0.8\linewidth]{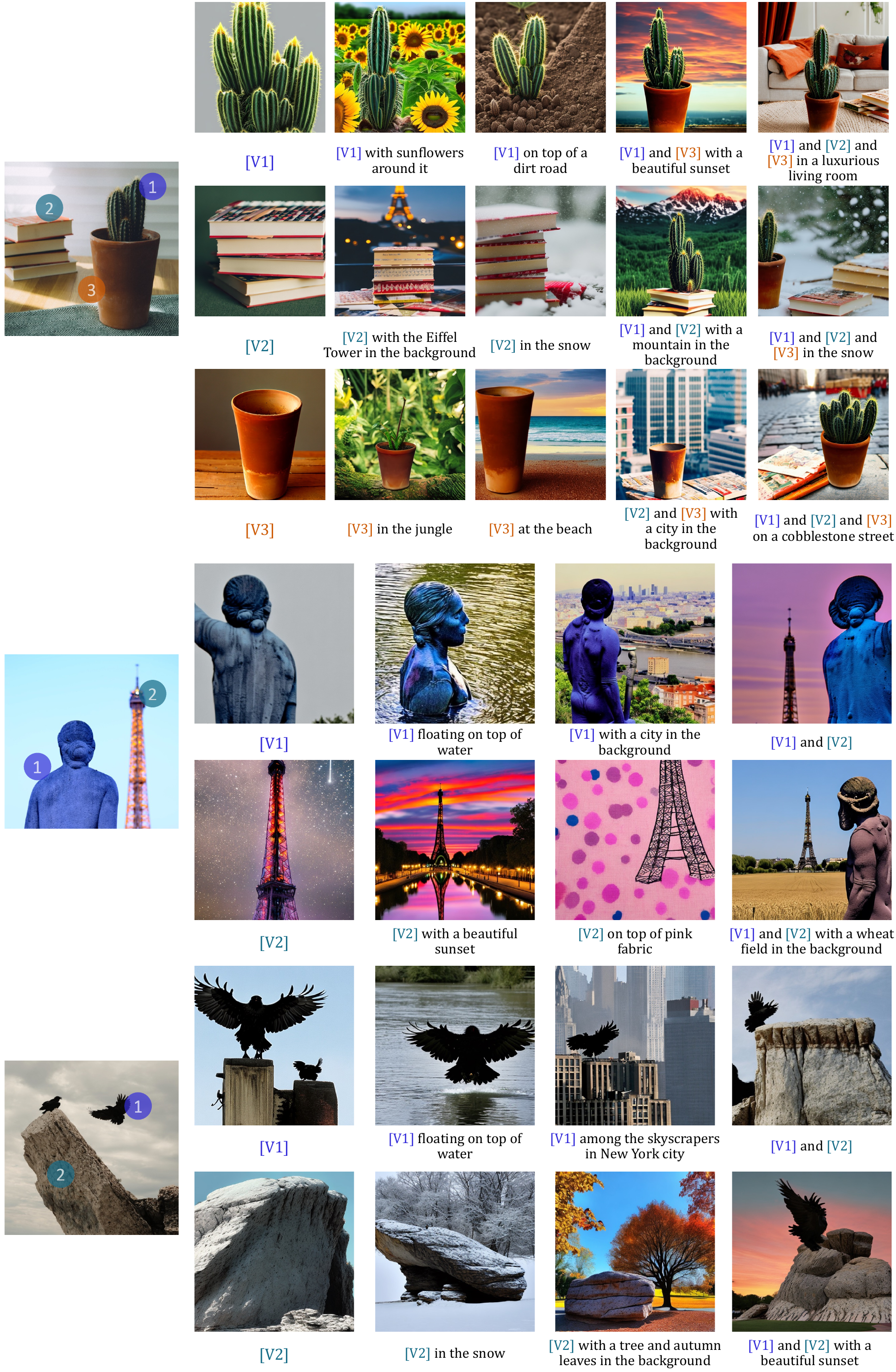}
    \caption{\textbf{Additional generated results of \modelname (part 1).}}
    \label{fig:supp-ours-p1}
\end{figure*}

\begin{figure*}
    \centering
    \includegraphics[width=0.9\linewidth]{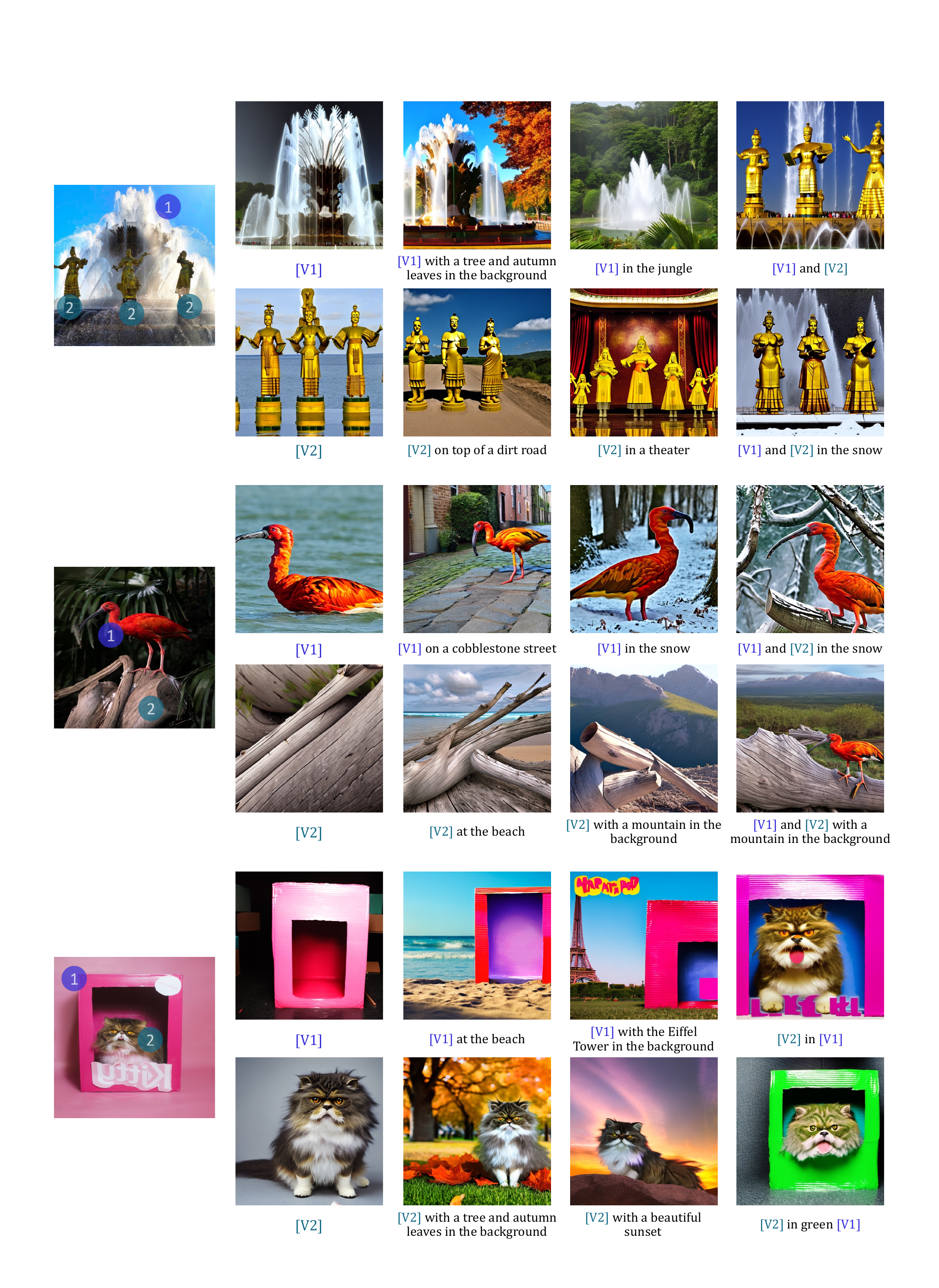}
    \caption{\textbf{Additional generated results of \modelname (part 2).}}
    \label{fig:supp-ours-p2}
\end{figure*}

As a supplement to the main paper, we provide additional comparison results between \modelname and Bas$^\dag$ in~\cref{fig:supp-compare}. Furthermore, we present a broader range of generation examples from \modelname in~\cref{fig:supp-ours-p1,fig:supp-ours-p2}, showcasing individual, compositional, and text-guided generation results. These results fully demonstrate the effectiveness of \modelname in the UCE problem.

\section{Human Interaction with SAM}
\begin{figure}[t]
    \centering
    \includegraphics[width=0.5\linewidth]{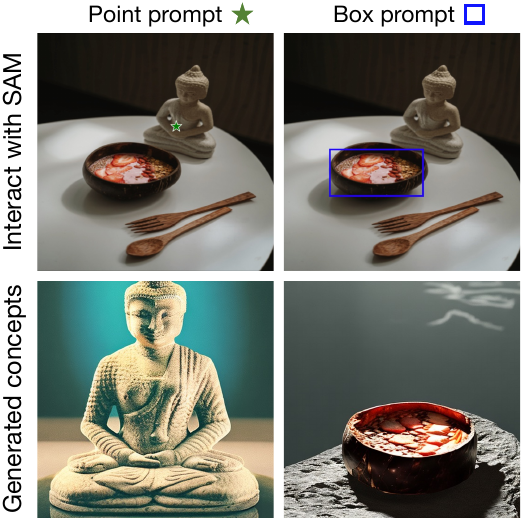}
    \caption{\textbf{Human interaction with SAM through point or box prompts}. By leveraging SAM, the human-desired entity in the given image can be explicitly specified and the corresponding concept can be effectively extracted and learned by \modelname.}
    \label{fig:interact_sam}
\end{figure}

\rvs{Although our task does not demand annotated concept masks, we can also seamlessly incorporate SAM~\cite{kirillov2023segment} into our model to enable interactive concept extraction. We showcase human interaction with SAM through point or box prompts in~\cref{fig:interact_sam}. This experiment demonstrates that our model can be seamlessly integrated with SAM in practice, enabling human interaction in the concept extraction process through explicit point or box prompts.}

\section{Unsatisfactory Cases}
In our analysis, we have noticed two limitations of \modelname. 
The first limitation is its difficulty in distinguishing instances from the same semantic category. The second limitation is its struggle to accurately learn concepts for instances with a relatively small occurrence. We have discussed these limitations in Sec.~\textcolor{red}{5} in the main paper. 

\begin{figure*}[t]
    \centering
    \includegraphics[width=\linewidth]{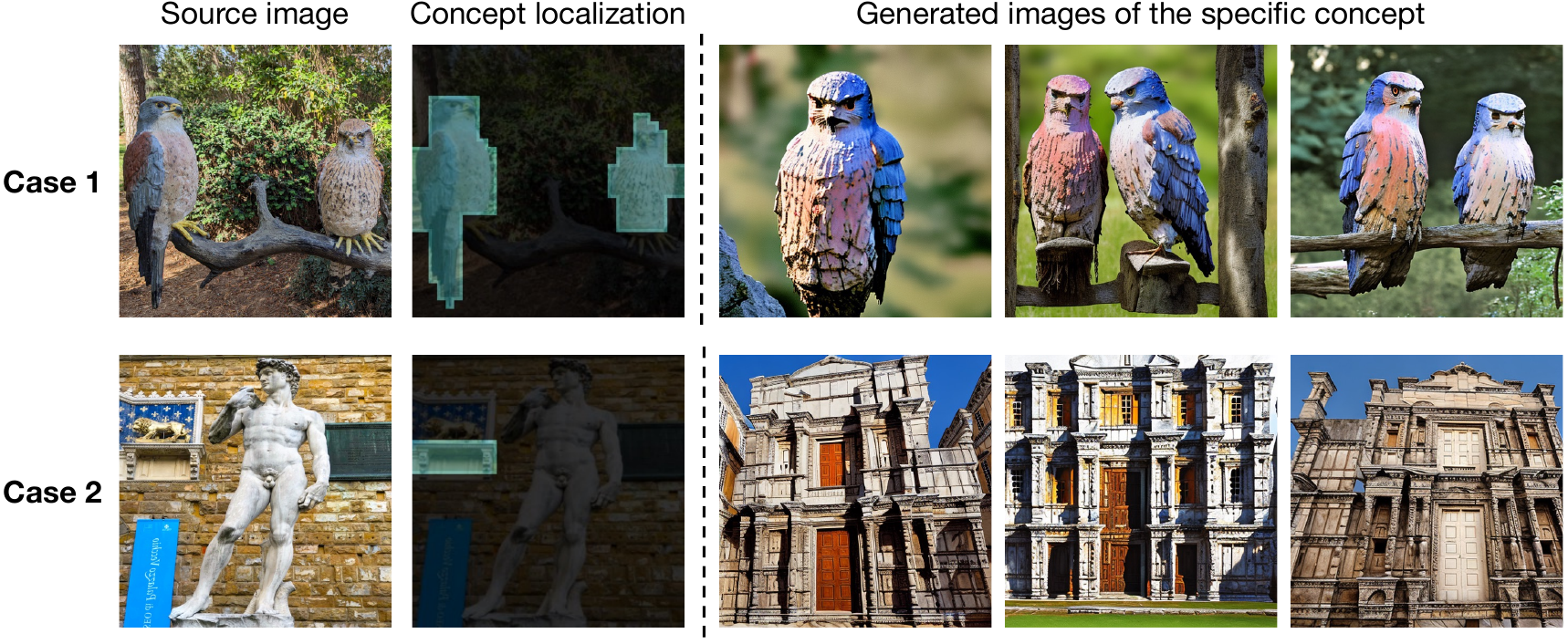}
    \caption{\textbf{Unsatisfactory cases.} \textbf{Case 1}: both instances belong to the same category (bird). \textbf{Case 2}:  the discovered concept has a small occurrence in the image. In the figure, we only visualize the concept of interest, while the other concepts are not plotted. }
    \label{fig:supp-fail}
\end{figure*}

To further illustrate these limitations, we provide examples of unsatisfactory cases in~\cref{fig:supp-fail}.
\crhsz{\textbf{The first case} arises because similar patches of instances from the same category (such as bird wings) exhibit close distributions in self-attention maps, leading to early grouping in the pre-clustering phase. As a result, we localize the two birds as a single concept, which may affect the number of instances shown in the generated image. We considered including spatial bias to address this, but we found it might impede the normal grouping of complete concepts. Therefore, we opt for our current approach.
It is important to note that this limitation does not significantly impact the overall generation quality. Our model preserves the integrity of complete concepts, ensuring they reflect bird characteristics rather than generating creatures with multiple heads. Despite the segmentation containing multiple instances, the model can still generate a single instance, as shown in row~1 column~3.}
\textbf{The second case} is caused by the limited occurrence of the target concept. Due to the low resolution (64$\tiny{\times}$64) of the latent space, the small region captures limited information, making it challenging to train an accurate concept representation. As a result, the small ``marble pedestal'' is learned through the diffusion process and is transformed into a representation resembling a ``marble church''. 
We believe that future research will aim to address these limitations.

\section{Broader Impact}
This paper presents a convenient and efficient concept extraction method that does not require external manual intervention. The extracted concepts can be used to generate new images. On one hand, this greatly facilitates the effortless disentanglement of instances from an image and enables the generation of personalized images. It even allows for the creation of a vast concept library by processing a large batch of images, which can be archived for the use of swift generation. On the other hand, however, this technology can also be meticulously exploited to handle and generate sensitive images, such as violent, pornographic, or privacy-compromising content. Moreover, due to its unsupervised nature, organizations with massive image datasets can easily perform concept extraction and build extensive concept libraries, which may include a substantial amount of harmful or sensitive content. We believe it is crucial to continuously observe and regulate this technology to ensure its responsible use in the future.

\paragraph{Limitation} 
\modelname has the following limitations that remain to be addressed in future research.
The first limitation is related to processing images containing multiple instances from the same semantic category, such as two instances of a bird. In this case, self-attention correspondence struggles to disentangle these instances and tends to identify them as a single concept, rather than recognizing them as separate instances.
The second limitation is that certain concepts with a relatively small occurrence in the image may be discovered. This can result in poor concept learning due to the lack of sufficient information for reconstructing the concept in the latent space with a resolution of 64$\tiny{\times}$64.
\crhsz{The third limitation is that our model requires a certain level of input image quality. In the future, it can be further enhanced to robustly handle uncurated natural data, making it more applicable to real-world scenarios.}

\newpage

\end{document}